%% file: main.tex
\documentclass[10pt,twocolumn,letterpaper]{article}

\usepackage{iccv}

\definecolor{iccvblue}{rgb}{0.21,0.49,0.74}
\usepackage[pagebackref,breaklinks,colorlinks,allcolors=iccvblue]{hyperref}

\input{preamble}  
\usepackage{colortbl}
\usepackage{multirow}
\usepackage{hyperref}
\usepackage[many]{tcolorbox}
\usepackage{xcolor}
\usepackage{xfp}

\title{On Large Multimodal Models as Open-World Image Classifiers}

\author{%
  Alessandro Conti\textsuperscript{1,}\thanks{Correspondence to: \texttt{alessandro.conti-1@unitn.it}.}\quad
  Massimiliano Mancini\textsuperscript{1}\quad 
  Enrico Fini\textsuperscript{2,}\thanks{Enrico Fini is currently at Apple}\quad \\
  Yiming Wang\textsuperscript{3}\quad
  Paolo Rota\textsuperscript{1}\quad \vspace{.1em}
  Elisa Ricci\textsuperscript{1,3} \vspace{1em} \\
  \textsuperscript{1}University of Trento\quad
  \textsuperscript{2}Independent researcher\quad\textsuperscript{3}Fondazione Bruno Kessler 
}

\begin{document}
\maketitle
\input{sec/0_abstract}    
\input{sec/1_intro}
\input{sec/2_related}
\input{sec/3_protocol}
\input{sec/4_analyses}
\input{sec/5_conclusions}

{
    \small
    \bibliographystyle{ieeenat_fullname}
    \bibliography{main}
}

\newpage
\input{sec/6_supmat}

\end{document}

%% file: preamble.tex
\definecolor{DrawioGreen}{RGB}{213,232,212}
\definecolor{DrawioRed}{RGB}{248,206,204}
\definecolor{skyblue}{RGB}{135, 206, 235}
\definecolor{lightcoral}{RGB}{240, 128, 128}
\definecolor{ultraviolet}{HTML}{8365BA}

\newcommand{\corrspec}{\textcolor{blue}{correct and specific}}
\newcommand{\corrgen}{\textcolor{skyblue}{correct but generic}}
\newcommand{\wrongspec}{\textcolor{lightcoral}{wrong but specific}}
\newcommand{\wronggen}{\textcolor{red}{wrong and generic}}

\newcommand{\ccc}[1]{%
  \ifnum\fpeval{#1 < -20}=1
    \cellcolor{DrawioRed!150}%
  \else\ifnum\fpeval{#1 < -10}=1
    \cellcolor{DrawioRed!100}%
  \else\ifnum\fpeval{#1 < 0}=1
    \cellcolor{DrawioRed!50}%
  \else\ifnum\fpeval{#1 > 20}=1
    \cellcolor{DrawioGreen!150}%
  \else\ifnum\fpeval{#1 > 10}=1
    \cellcolor{DrawioGreen!100}%
  \else\ifnum\fpeval{#1 > 0}=1
    \cellcolor{DrawioGreen!50}%
  \else
    \cellcolor{white}%
  \fi\fi\fi\fi\fi\fi
  #1%
}

\newcommand{\inlineColorbox}[2]{\begingroup\setlength{\fboxsep}{1pt}\colorbox{#1}{\hspace*{2pt}\vphantom{Ay}#2\hspace*{2pt}}\endgroup}

%% file: sec/0_abstract.tex
\begin{abstract}
Traditional image classification requires a predefined list of semantic categories.
In contrast, Large Multimodal Models (LMMs) can sidestep this requirement by classifying images directly using natural language (\eg, answering the prompt \textit{``What is the main object in the image?''}).
Despite this remarkable capability, most existing studies on LMM classification performance are surprisingly limited in scope, often assuming a closed-world setting with a predefined set of categories.
In this work, we address this gap by thoroughly evaluating LMM classification performance in a truly open-world setting.
We first formalize the task and introduce an evaluation protocol, defining various metrics to assess the alignment between predicted and ground truth classes.
We then evaluate 13 models across 10 benchmarks, encompassing prototypical, non-prototypical, fine-grained, and very fine-grained classes, demonstrating the challenges LMMs face in this task.
Further analyses based on the proposed metrics reveal the types of errors LMMs make, highlighting challenges related to granularity and fine-grained capabilities, showing how tailored prompting and reasoning can alleviate them.
Code is available at \url{https://github.com/altndrr/lmms-owc}.
\end{abstract}

%% file: sec/1_intro.tex
\section{Introduction}
\label{sec:intro}
{Image classification aims to assign a label 
to an image. }
This widely studied task relies on a key assumption: the categories are fixed and known in advance, a setting known as the \textit{closed world}.
However, the latter is often restrictive in real-world applications where new categories can emerge, requiring to expand the label set~\cite{belouadah2021ILsurvey}, recognize unseen concepts~\cite{geng2020open-set-survey}, or both~\cite{bendale2015towards-open-world}.
Despite its limitations, this assumption has historically been useful, enabling supervised training and straightforward evaluation on labeled datasets.
With the rise of {Large Multimodal Models (LMMs)~\cite{liu2023llava,li2023blip2,bai2023qwen}} processing images and text, this constraint is no longer necessary.
Instead of choosing from a fixed list, LMMs can answer open-ended prompts such as \textit{“What is the object in the image?”}, recognizing virtually {any} semantic concept.
From this perspective, closed-world classification is an artificial limitation that restricts a model’s expressive capabilities rather than reflecting its true potential.

\begin{figure}
\includegraphics[width=\linewidth]{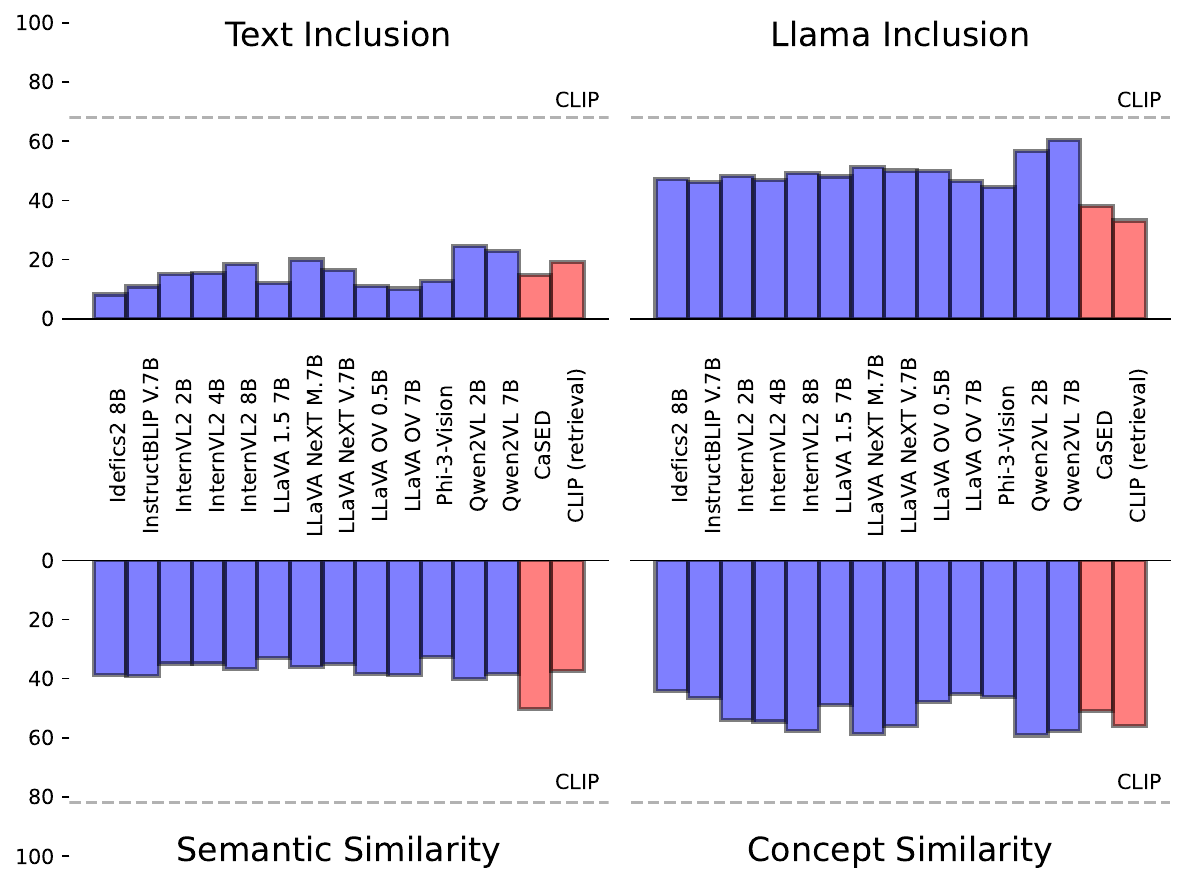}
\caption{
    We extensively test 13 Large Multimodal Models (LMMs) for Open-World (OW) classification on 10 datasets using four evaluation metrics.
    We show that \inlineColorbox{blue!20}{LMMs} outperform \inlineColorbox{red!20}{contrastive-based approaches} in OW (CaSED~\cite{conti2023vocabulary}, and CLIP~\cite{radford2021clip} with image-to-text retrieval) but still lag behind closed-world models with fixed categories (CLIP~\cite{radford2021clip}, dashed line).
}
\vspace{-10pt}
\label{fig:teaser}
\end{figure}

While some studies have explored classification with LMMs, they have either focused on the closed-world setting~\cite{liu2024revisiting} or relied on limited metrics to assess performance: checking whether the ground truth label appears in the model’s prediction~\cite{zhang2024visually}.
However, this metric provides a limited view of classification performance.
It fails to account for alternative correct answers (\eg, \textit{sofa} instead of \textit{couch}), while also overlooking real mistakes (\eg, confusing \textit{can} with \textit{trash can}).
Evaluating models in the open world presents additional challenges, as predictions may differ in granularity (\eg, \textit{dog} \vs \textit{pug}), or conflict with annotation ambiguities (\eg, \textit{bedroom} vs \textit{bed}).
These issues highlight the need for a more comprehensive evaluation framework to assess the open-world capabilities of LMMs.

In this work, we address this gap by formalizing the Open-World (OW) classification task and introducing four complementary metrics: (i) text inclusion~\cite{zhang2024visually}, evaluating string matching, (ii) Llama inclusion which leverages Llama~\cite{grattafiori2024llama}, to distinguish good and bad responses, as in LLM-as-a-judge~\cite{zheng2023judging}, (iii) semantic similarity~\cite{conti2023vocabulary} between text embeddings of predictions and ground truth, and (iv) concept similarity, doing the same at the level of sentence parts.
Using these metrics, we evaluate 13 models across 10 benchmarks spanning different levels of granularity, from prototypical, coarse categories (\eg, Caltech101~\cite{fei2004caltech101}) to fine-grained (\eg, Flowers102~\cite{nilsback2008flowers}), very fine-grained (\eg, Stanford Cars~\cite{krause20133cars}), and non-prototypical ones (\eg, DTD~\cite{cimpoi2014dtd}).
Our results (aggregated in \cref{fig:teaser}) show that these models often predict semantically related concepts (even better than previous, contrastive-based alternatives~\cite{conti2023vocabulary}), demystifying the skepticism against LMMs on OW classification.
However, LMMs also make notable errors, being far from closed-world baselines.

While the challenging nature of the problem makes analyzing the severity of the mistakes hard, we find that different behaviors on different metrics can pinpoint sources of errors.
In particular, mismatches in concept similarity and Llama inclusion uncover errors in the granularity of the predictions (\ie, correct but generic) or very similar categories easy to confuse (\ie, wrong but specific).
We show how the former can be addressed via tailored prompting, while the latter via implementing reasoning strategies.
Finally, we further analyze cases where the metrics identify a mistake and, using tagging models~\cite{huang2023ramplus}, check if predictions are incorrect only due to the single-label nature of the datasets.

\vspace{2pt}
\noindent\textbf{Contributions.} To summarize, our contributions are:
    \begin{itemize}
        \item We formalize a comprehensive evaluation protocol for the task of \textit{open-world classification} with LMMs, using 4 different metrics capturing both semantic and text alignment of predictions with the ground truth.
        \item We perform the first, large-scale assessment of LMMs on this task, using 13 models on 10 benchmarks, showing promising results yet multiple challenging cases.
        \item By combining the different metrics, we investigate the root of the models' mistakes, identifying various issues (\eg, wrong granularity, fine-grained discrimination, labeling ambiguities) and showing how changes in the models (\eg, prompts, reasoning) can reduce them.
        \item We use these results to draw conclusions on the source of errors that future research should account for when using these models, releasing our evaluation suite to encourage future research efforts on addressing them.
    \end{itemize}

%% file: sec/2_related.tex
\section{Related Work}
\label{sec:related}

\noindent\textbf{Large Multimodal Models.}
While early large vision-language models aligned visual and text representation in a shared embedding space~\cite{radford2021clip,zhai2023siglip}, there has been an increasing effort in developing generative multimodal models~\cite{alayrac2022flamingo,li2023blip2,zhu2024minigpt4,singh2022flava}.
These models process an input image and text generating either a text~\cite{li2023blip2,liu2023llava} or a multimodal~\cite{koh2023fromage,zheng2023minigpt5} output.
While these models share common components (\eg, visual and text encoders, text decoders) they differ in the specific strategies for modality alignment  (\eg, MLP projector~\cite{liu2023llava}, Q-former~\cite{li2023blip2}), pretraining (\eg, autoregressive~\cite{liu2023llava}, alignment~\cite{li2023blip2}), fine-tuning (\eg, supervised, instruction tuning) but also data source (\eg, web data~\cite{laurenccon2025idefics}, textbook-style~\cite{abdin2024phi}) and structure (\eg, captioning~\cite{dai2023instructblip}, interleaved image-text~\cite{laurenccon2025idefics}).
In this work, we do not aim to introduce a new LMM or novel methodologies for building LMMs.
Instead, we focus on evaluating how these models perform in OW classification, testing 13 different models belonging to 8 different families~\cite{laurenccon2025idefics,chen2024internvl,dai2023instructblip,liu2023llava,li2024llavaNext,li2024llava-ov,abdin2024phi,bai2023qwen}, covering multiple architectural, data, and design choices.

\noindent\textbf{Classification with LMMs.} Multiple works designed benchmarks to test the general capabilities~\cite{li2024seed,liu2024mmbench,li2024mvbench}, or shortcomings~\cite{hsieh2023sugarcrepe,yuksekgonuland,li2023evaluating,zhou2023analyzing} of LMMs.
The most closely related works to ours are \cite{yue2024object, zhang2024visually, liu2024revisiting}, investigating their classification performance.
Yue \textit{et al.}~\cite{yue2024object} developed an approach exploiting the next token prediction probability of an LMM, reporting results on multi-label recognition.
Zhang \textit{et al.}~\cite{zhang2024visually} tested multiple LMMs on both closed-world and OW settings, showing how data influences their performance and that generative LMMs underperform their contrastive counterpart.
This latter finding is challenged by Liu \textit{et al.}~\cite{liu2024revisiting}, who extended the analyses of \cite{zhang2024visually} to multiple datasets and more recent models.
However, \cite{liu2024revisiting} focused on closed-world classification, while \cite{zhang2024visually} limited the analyses on OW to 4 datasets and a single metric (\ie, text inclusion).
In this work, we expand existing analyses in OW classification with LMMs, providing the largest study up-to-date in terms of datasets (10) and models (13).
We also analyze the performance of LMMs according to four different metrics, capturing complementary aspects.
Moreover, we use these metrics to analyze LMM mistakes in this scenario.

\noindent\textbf{Analyzing model failures.} There has been a growing interest in studying what type of mistakes models make.
For instance, works on \textit{failure modes} detection studied how to identify slices of data on which models underperform~\cite{eyubogludomino,singla2021understanding,yenamandra2023facts} and, through the use of LMMs, these slices can be also interpreted via natural language~\cite{kim2024discovering,dunlap2024describing,csurka2024could}.
Other works examined the models' mistakes on specific datasets, to understand what prevents them from achieving perfect performance and to provide guidelines for future works.
This has been the case for ImageNet~\cite{russakovsky2015imagenet}, where previous studies discovered problems linked to spurious correlations~\cite{moayeri2022hard,singlasalient}, fine-grained discrimination~\cite{vasudevan2022does,peychev2023automated}, but also labeling itself~\cite{beyer2020we}.
Our work is similar to this latter trend, as we want to investigate what type of mistakes LMMs make when classifying images in the OW.
We aim for our findings to serve as a foundation for future research focused on improving the performance of LMMs in this challenging task.

%% file: sec/3_protocol.tex
\section{Benchmarking LMMs in OW Classification}
\label{sec:protocol}

In this section, we first formalize the setting of OW classification with LMMs, clarifying its goal and terminology w.r.t.~related works (\cref{sec:task}).
We then discuss how to evaluate performance in this setting, describing the different metrics and what they capture (\cref{sec:metrics}).
Last, we provide details on the datasets and models considered in our analyses (\cref{sec:benchmark}) before showing their results (\cref{sec:results}).

\subsection{Preliminaries}
\label{sec:task}
\noindent\textbf{Classification with LMMs.}~Let us define an LMM as a function $f_\mathtt{LMM}$ generating a text output $y$ in the space $\mathcal{T}$ given an image $x$ in the space $\mathcal{X}$ and a text query $q\in\mathcal{T}$, \ie, $f_\mathtt{LMM}:\mathcal{X}\times\mathcal{T}\rightarrow\mathcal{T}$.
To perform classification with LMMs, the query $q$ contains a prompt of the type \textit{``What type of object is in this image?''} and we expect the output $y$ to be a semantic class $\mathcal{Y}\subset\mathcal{T}$.
In the case of closed-world classification, we have a predefined list $\mathcal{C}$ of classes and we modify $q$ by specifying the set $\mathcal{C}$ (\eg, via a multi-choice question).
In OW we let the LMM predict naturally on its original output space $\mathcal{T}$, without any constraint.
As a consequence,  the model can pick from the set $\mathcal{Y}$ of \textit{all} possible semantic concepts, with $\mathcal{C}\subset\mathcal{Y}$ and $|\mathcal{C}|\ll|\mathcal{Y}|$.

\vspace{2pt}
\noindent\textbf{Relationships with prior problem definitions.}
While we followed \cite{zhang2024visually} and used \textit{open-world} to define this setting, the term can be ambiguous.
The traditional definition of OW recognition~\cite{bendale2015towards-open-world} refers to a different problem, where a model trained to recognize $\mathcal{C}$ classes should recognize whether an instance belongs to an unknown one $u\notin \mathcal{C}$ and learn to recognize $u$.
Other works refer to this task as \textit{vocabulary-free} classification~\cite{conti2023vocabulary} due to the absence of a predefined vocabulary, \textit{open-ended} recognition~\cite{yu2024towards} due to the lack of constraints, or avoided any specific terminology in the context of multi-label recognition~\cite{yue2024object}.
While these different definitions closely relate to each other, we follow \cite{zhang2024visually}, clarifying that OW here refers only to the lack of constraints in the output space of the LMMs.

\subsection{Metrics}
\label{sec:metrics}
Evaluating open-world recognition with LMMs is challenging as, even if we have a ground truth, we have no guarantee that the model will output the same name when correct (\eg, \textit{sofa} vs \textit{couch}), especially as the model may produce an undesired wordy output (\eg, \textit{the object in the image is a sofa}).
These potential variations ask for specific evaluation criteria, accounting for different types of (mis)alignment between the prediction and the ground truth.
Below, we describe the four metrics we consider for this task.

\noindent\textbf{Text inclusion (TI).} This metric, adopted in \cite{zhang2024visually}, refers to whether the ground truth is contained in the model's prediction.
Specifically, let us define as $y$ the ground truth and as $\hat{y}$ the model's prediction.
Text inclusion score is defined as:
\begin{equation}
    \text{TI}(y, \hat{y}) =\left\{     
    \begin{matrix}
            1 & \text{if}\, y \subseteq \hat{y}, \\
            0 & \text{otherwise} \\
    \end{matrix}
    \right.
\end{equation}
where, in this context, $\subseteq$ refers to string inclusion.
This metric assesses whether the predictions strictly adhere to the ground truth label but over-penalizes whether the two are semantically coherent (\eg, the prediction \textit{labrador} would be considered wrong for the label \textit{labrador dog}).

\input{tab/grouped_all}

\vspace{2pt}
\noindent\textbf{Llama inclusion (LI).} Differently from TI, this metric evaluates whether the prediction aligns with the ground truth label based on a Large Language Model (LLM) internal knowledge.
Specifically, we employ Llama 3.2 3B~\cite{touvron2023llama} and report the prompt we use in the Supp. Mat. (see~\ref{sec:supp-metrics}).
The score is 0 or 1, depending on the LLM's answer.
This is similar to methods that use LLM/LMMs-as-a-judge~\cite{zheng2023judging,chen2024mllmasajudge}, but is specifically adapted to OW classification.

\vspace{2pt}
\noindent\textbf{Semantic similarity (SS).} 
Unlike previous metrics that assess alignment with the ground truth in a binary manner, SS captures the degree of semantic similarity on a continuous scale between 0 and 1.
To achieve this, we employ a semantic similarity metric.
Following~\cite{conti2023vocabulary}, we define similarity as $\langle g_\mathtt{emb}(\hat{y}), g_\mathtt{emb}(y) \rangle$, where $g_\mathtt{emb}$ is a text embedding function, and $\langle\cdot, \cdot \rangle$ denotes cosine similarity.
As in~\cite{conti2023vocabulary}, we use Sentence-BERT~\cite{reimers2019sentence} for computing embeddings.

\vspace{2pt}
\noindent\textbf{Concept similarity (CS).} By considering the prediction as a whole, the semantic similarity previously defined ignores whether parts of the sentence (\eg, \textit{elephant}) are closer to the ground truth (\eg, \textit{animal}) than the sentence as a whole (\eg, \textit{a photo of an elephant in the room}).
To address this, we consider CS as an additional metric, defining it as:
\begin{equation}
    \label{eq:cs}
    \max_{p\in \mathtt{split}(\hat{y})} \langle g_\mathtt{emb}(p), g_\mathtt{emb}(y) \rangle
\end{equation}
where $\mathtt{split}$ is a sentence splitting procedure that, in our case, is implemented via spaCy~\footnote{We use the model available at \url{https://spacy.io/models/en\#en_core_web_lg}}.

\subsection{Dataset and Models}
\label{sec:benchmark}

\noindent\textbf{Datasets.} Following previous works~\cite{conti2023vocabulary,shu2022tpt,zhou2022learning}, we analyze four different challenges: coarse-grained (or prototypical), non-prototypical, fine-grained, and very fine-grained classification.
For the \textbf{prototypical} classification, we include standard benchmarks such as Caltech101~\cite{fei2004caltech101} for objects and SUN397~\cite{xiao2010sun} for places.
The \textbf{non-prototypical} set comprises datasets that either lack nouns or involve non-standard domains.
This includes DTD~\cite{cimpoi2014dtd} (textures), UCF101~\cite{soomro2012ucf101} (actions), and EuroSAT~\cite{helber2019eurosat} (satellite images).
The \textbf{fine-grained} set consists of datasets where classes belong to a shared superclass and/or are challenging to distinguish.
These include Flowers102~\cite{nilsback2008flowers} (flowers), Food101~\cite{bossard2014food} (food), and OxfordPets~\cite{parkhi2012pets} (animals).
Finally, the \textbf{very fine-grained} set comprises datasets where categories are not only within the same subclass but also highly difficult to differentiate.
This includes StanfordCars~\cite{krause20133cars}, where labels specify car brands, models, and years of production, and FGVCAircraft~\cite{maji2013aircraft}, which categorizes aircraft models.
More details are in the Supp. Mat. (see~\ref{sec:supp-dm}).

\noindent\textbf{Models.}
We perform our evaluation considering  state-of-the-art
LMMs of 8 types, including {Idefics2}~\cite{laurenccon2025idefics}, InstructBLIP \cite{dai2023instructblip}, InternVL2 \cite{chen2024internvl,chen2024far}, {LLaVA-1.5} \cite{liu2023llava}, LLaVA-NeXT \cite{li2024llava-ov}, LLaVa-OV \cite{li2024llava-ov}, Phi-3-Vision \cite{abdin2024phi}, Qwen2VL \cite{wang2024qwen2}.
We choose these models as they are publicly available and widely adopted by the community.
These models encompass different design choices such as the vision encoder (\eg, CLIP~\cite{radford2021clip}, SigLIP~\cite{zhai2023siglip}, BLIP-2~\cite{li2023blip2}), language model (\eg, Mistral~\cite{jiang2023mistral7b}, Vicuna~\cite{chiang2023vicuna}, Qwen2~\cite{wang2024qwen2}), data types (\eg, instruction following, multilingual, textbook-based), and pretraining strategies (\eg, single vs multi-stage).
Unless otherwise stated, we query the model with the same prompt of \cite{zhang2024visually}, \ie, \textit{``What type of object is in this image?''}, letting the models perform unconstrained generation.
We report a summary of the models and their differences in the Supp. Mat. (see~\ref{sec:supp-dm}).
    
\noindent\textbf{References.} As a reference, we consider baselines based on contrastive vision-language models.
Specifically, we report results using CLIP~\cite{radford2021clip} and SigLIP~\cite{zhai2023siglip} in the closed-world setting, where the models have access to the list of target class names.
Additionally, we include two baselines that adapt CLIP to the OW setting by formulating image classification as a retrieval task.
The first retrieves the closest caption from a predefined database, while the second, CaSED~\cite{conti2023vocabulary}, leverages retrieved captions to generate a list of candidate classes for the final prediction.
For both baselines, we use the same retrieval database as in~\cite{conti2023vocabulary}.

\begin{figure*}
\centering
\begin{tabular}{llllll}
{\includegraphics[width=2.5cm]{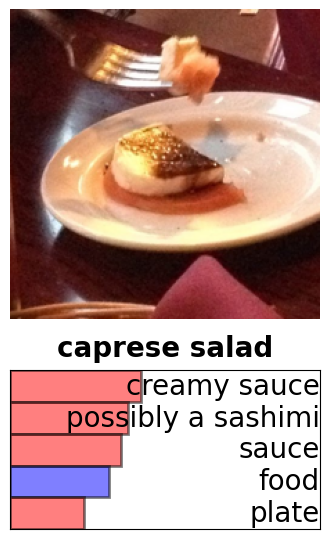}} &
{\includegraphics[width=2.5cm]{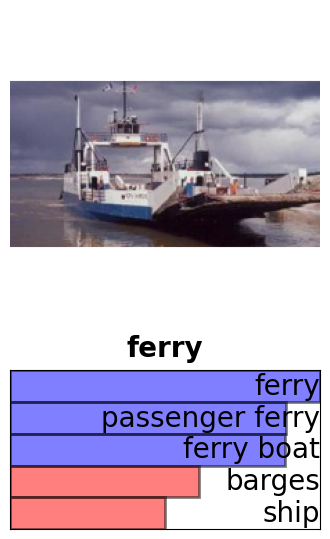}} &
{\includegraphics[width=2.5cm]{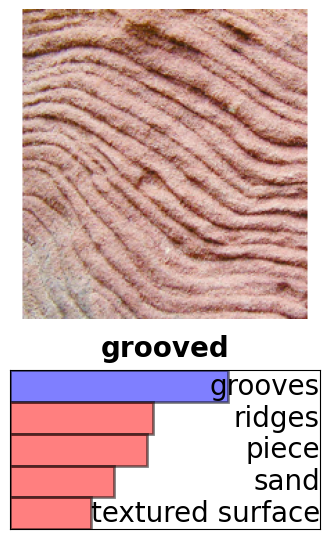}} &
{\includegraphics[width=2.5cm]{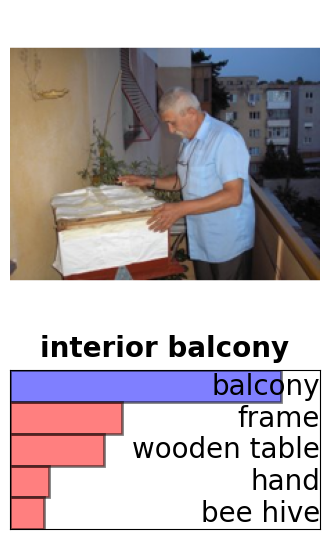}} &
{\includegraphics[width=2.5cm]{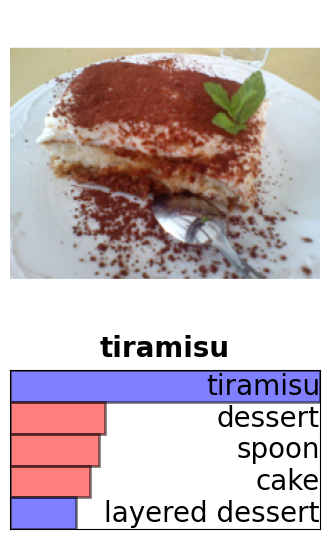}} &
{\includegraphics[width=2.5cm]{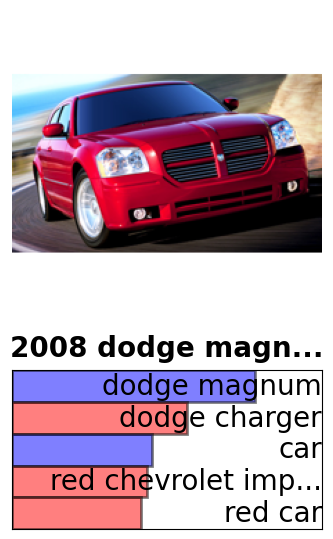}} \\
\end{tabular}
\caption{
    Per-image examples of model predictions.
    \textbf{Bold} indicates the ground truth class names.
    For visualization purposes, we show only the predictions with the highest/lowest concept similarity.
    \inlineColorbox{blue!50}{Blue} and \inlineColorbox{red!50}{red} indicate positive and negative Llama inclusion values.}
\label{fig:metrics_per_image_qualitative}
\end{figure*}

\begin{figure*}
\centering
\begin{tabular}{llllll}
{\includegraphics[width=4.0cm]{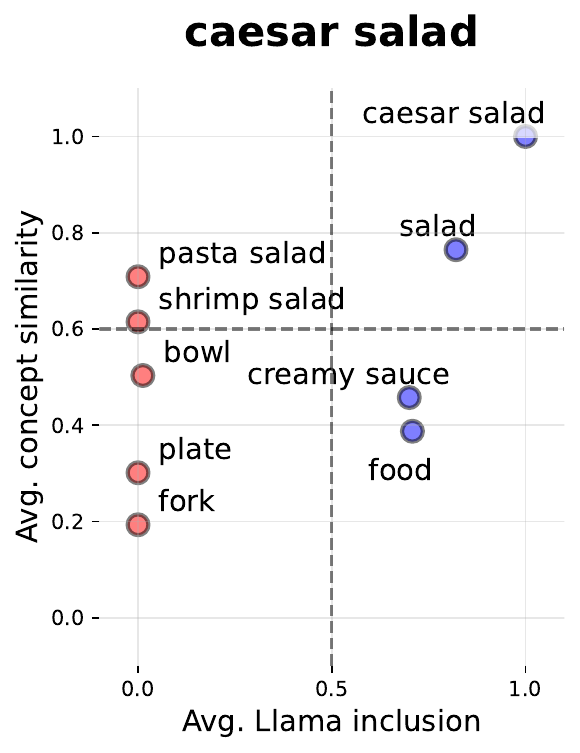}} &
{\includegraphics[width=4.0cm]{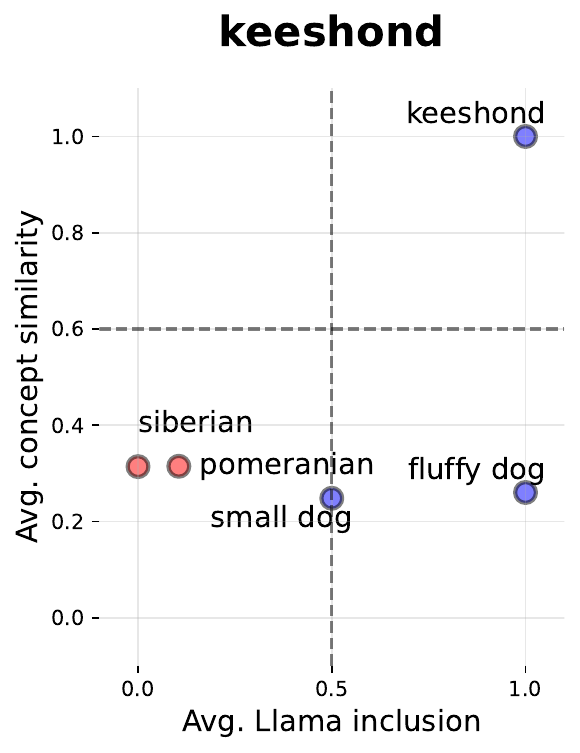}} &
{\includegraphics[width=4.0cm]{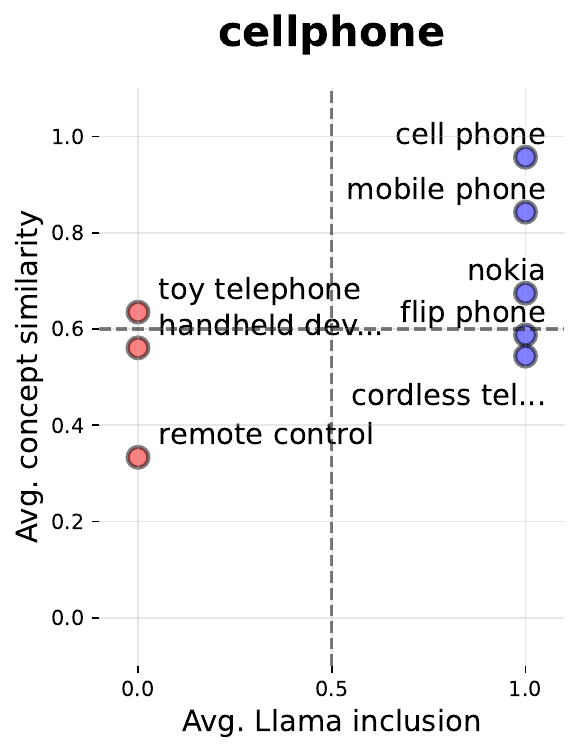}} &
{\includegraphics[width=4.0cm]{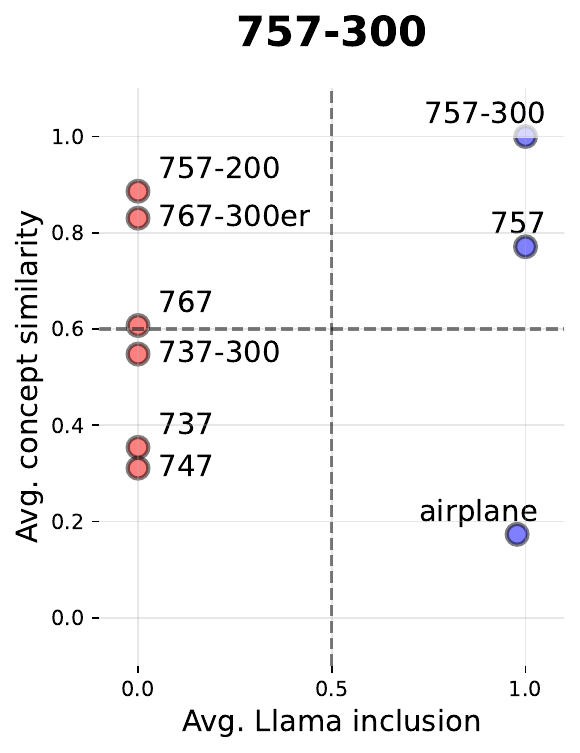}} &
\end{tabular}
\vspace{-10pt}
\caption{
    Per-class examples of model predictions.
    \textbf{Bold} indicates the ground truth class names.
    On the x-axis we report the average LI, and on the y-axis the average CS.
    For visualization purposes, we show the most frequent concepts predicted for each quadrant.}
\vspace{-14pt}
\label{fig:metrics_per_class_qualitative}
\end{figure*}

\subsection{Are LMMs Good at OW Classification?}
\label{sec:results}

In this section, we analyze the performance of LMMs in an OW setting, with results summarized in~\cref{tab:grouped_all} by dataset groups with per-dataset results in the Supp. Mat., see~\ref{sec:supp-results}).

\noindent\textbf{Prototypical classification.} LMMs perform best on prototypical classes, with high scores on inclusion and similarity metrics.
They consistently outperform CaSED and CLIP retrieval on inclusion metrics and are generally comparable or superior on similarity scores.

\noindent\textbf{Non-prototypical classification.} Performance drops significantly, with the highest LI score at 46.8, nearly 15 points lower than closed-world CLIP.
Predictions are also less semantically indicative of the target class, with an average CS of 49.3, much lower than the prototypical case (69.2).

\noindent\textbf{Fine-grained classification.} Greater variation is observed among different models, ranging from 41.7 to 63.4 in concept similarity.
In this group, LMM predictions are slightly less accurate than those of CaSED and CLIP retrieval.

\noindent\textbf{Very fine-grained classification.} Many models achieve a TI of 0.0, except for Qwen2VL~2B, which scores 12.9 due to the exceptional performance of FGVAircraft (25.6 vs 4.6 for the second-best model).
Most LMMs underperform CLIP retrieval in CS, suggesting an issue due to granularity.

\noindent\textbf{Overall trends.}
Across ten datasets, CLIP retrieval outperforms 9/13 models in TI, but LMMs consistently achieve higher Llama inclusion scores.
CaSED ranks highest in semantic similarity due to its concise responses, while CLIP retrieval remains competitive.
These results confirm insights from previous works, such as the influence of data exposure on coarse-grained categories \cite{zhang2024visually,liu2024revisiting}.
Additionally, stronger language models (\eg, Mistral, Qwen) tend to yield better results than weaker counterparts (\eg, Vicuna).
LMMs generally outperform contrastive models in OW classification, leading in 11/16 metric/groups.
However, top-performing models in one metric may struggle in others--for example, Qwen2VL~7B excels in LI on very fine-grained datasets, while InternVL2~8B and LLaVA-OV (Qwen2~7B) show different strengths in prototypical classification: \eg, +21.2 LI of the first on the second but -7.9~SS.

While these results are promising, there is still a large gap with closed-world models, \ie, CLIP~\cite{radford2021clip}, SigLIP~\cite{zhai2023siglip}.
In the next sections, we further explore what the metrics capture, to better understand OW predictions.

%% file: tab/grouped_all.tex
\begin{table*}
\centering
\resizebox{\linewidth}{!}{%
\begin{tabular}{lcccccccccccccccc}
\toprule
 & \multicolumn{4}{c}{\textbf{Prototypical}} & \multicolumn{4}{c}{\textbf{Non-prototypical}} & \multicolumn{4}{c}{\textbf{Fine-grained}} &  \multicolumn{4}{c}{\textbf{Very fine-grained}} \\
\cmidrule(lr){2-5} \cmidrule(lr){6-9} \cmidrule(lr){10-13} \cmidrule(lr){14-17}
\textbf{Model} & \textbf{TI} & \textbf{LI} & \textbf{SS} & \textbf{CS} & \textbf{TI} & \textbf{LI} & \textbf{SS} & \textbf{CS} & \textbf{TI} & \textbf{LI} & \textbf{SS} & \textbf{CS} & \textbf{TI} & \textbf{LI} & \textbf{SS} & \textbf{CS} \\
\midrule
\rowcolor{cyan!7} \textsc{Idefics2} \cite{laurenccon2025idefics} 8B & 30.8 & 52.7 & 54.5 & 63.1 & 3.7 & 27.9 & 35.4 & 41.3 & 3.0 & 49.9 & 38.0 & 41.7 & 0.0 & 67.0 & 29.6 & 33.6 \\
\textsc{InstructBLIP} \cite{dai2023instructblip} Vicuna 7B & 29.7 & 56.3 & 56.8 & 64.0 & 6.0 & 27.1 & 37.0 & 42.0 & 10.4 & 48.8 & 35.6 & 47.2 & 0.0 & 61.0 & 30.0 & 34.3 \\
\rowcolor{cyan!7} \textsc{InternVL2} \cite{chen2024internvl, chen2024far} 2B & 36.9 & 69.9 & 46.9 & 70.4 & 10.2 & 45.2 & 31.6 & 53.4 & 14.9 & 47.0 & 31.6 & 50.7 & 0.7 & 32.9 & 33.1 & 43.9 \\
\rowcolor{cyan!7} \textsc{InternVL2} \cite{chen2024internvl, chen2024far} 4B & 36.3 & 68.5 & 46.5 & 70.8 & 10.1 & 42.1 & 30.8 & 53.1 & 16.2 & 44.4 & 32.0 & 52.0 & 1.7 & 36.8 & 33.8 & 44.2 \\
\rowcolor{cyan!7} \textsc{InternVL2} \cite{chen2024internvl, chen2024far} 8B & 40.6 & 74.4 & 48.2 & 74.0 & 11.0 & 46.2 & 31.9 & {53.9} & 22.3 & 46.7 & 34.8 & 56.7 & 2.3 & 32.5 & 36.0 & 49.4 \\
\textsc{LLaVA-1.5} \cite{liu2023llava} 7B & 34.6 & 63.1 & 45.3 & 65.8 & 8.6 & 44.3 & 33.0 & 49.5 & 8.4 & 46.5 & 28.2 & 44.8 & 0.0 & 41.0 & 28.6 & 37.6 \\
\textsc{LLaVA-NeXT} \cite{li2024llavaNext} (Mistral 7B) & 41.7 & 73.9 & 45.9 & 74.3 & \textbf{11.3} & \textbf{46.8} & 31.2 & \textbf{54.4} & 26.8 & 43.7 & 35.3 & 60.1 & 1.4 & 47.2 & 34.2 & 46.9 \\
\textsc{LLaVA-NeXT} \cite{li2024llavaNext} (Vicuna 7B) & 39.5 & 72.8 & 46.2 & 73.2 & 10.6 & 45.9 & 31.1 & 54.2 & 16.9 & 44.5 & 32.2 & 53.2 & 1.3 & 42.2 & 34.5 & 46.1 \\
\rowcolor{cyan!7} \textsc{LLaVA-OV} \cite{li2024llava-ov} (Qwen2 0.5B) & 34.4 & 64.4 & 54.0 & 67.3 & 7.3 & 37.0 & 32.8 & 47.0 & 6.0 & 42.7 & 38.5 & 43.3 & 0.6 & 65.6 & 30.5 & 37.1 \\
\rowcolor{cyan!7} \textsc{LLaVA-OV} \cite{li2024llava-ov} (Qwen2 7B)  & 30.8 & 53.2 & 56.1 & 62.0 & 7.2 & 28.1 & 31.6 & 43.8 & 6.4 & 40.4 & 39.0 & 43.8 & 0.0 & 76.7 & 31.9 & 32.4 \\
\textsc{Phi-3-Vision} \cite{abdin2024phi} & 34.1 & 60.1 & 47.7 & 65.1 & 6.0 & 28.7 & 26.0 & 39.5 & 13.4 & 49.1 & 31.8 & 47.2 & 0.2 & 45.0 & 28.9 & 36.0 \\
\rowcolor{cyan!7} \textsc{Qwen2VL} \cite{wang2024qwen2} 2B & 44.9 & 77.8 & 52.2 & 74.7 & 7.8 & 34.3 & 27.7 & 42.7 & \textbf{35.7} & 62.5 & 40.7 & 63.4 & \textbf{12.9} & 60.7 & \textbf{45.1} & \textbf{62.3} \\
\rowcolor{cyan!7} \textsc{Qwen2VL} \cite{wang2024qwen2} 7B & \textbf{46.4} & \textbf{78.7} & 51.9 & \textbf{76.0} & 10.3 & 42.6 & 30.8 & 49.8 & {34.6} & \textbf{64.0} & 39.2 & 62.9 & 0.8 & \textbf{63.0} & 34.5 & 43.4 \\
\midrule
\rowcolor{pink!25} \multicolumn{17}{l}{\footnotesize \textit{Open-world baselines}} \\
\rowcolor{pink!25} \textsc{CaSED} \cite{conti2023vocabulary} & 24.5 & 46.3 & \textbf{58.9} & 59.8 & 5.4 & 18.6 & \textbf{41.8} & 42.4 & 27.4 & 46.6 & \textbf{60.7} & 61.7 & 0.7 & 47.1 & 38.5 & 38.5 \\
\rowcolor{pink!25} CLIP retrieval & 28.6 & 42.9 & 40.2 & 60.6 & 7.5 & 24.6 & 28.1 & 43.4 & 32.4 & 45.4 & 42.9 & \textbf{65.4} & 7.0 & 18.1 & 39.7 & 56.1 \\
\midrule
\rowcolor{gray!10} \multicolumn{17}{l}{\footnotesize \textcolor{gray}{\textit{Closed-world baselines}}} \\
\rowcolor{gray!10}\textcolor{gray}{CLIP \cite{radford2021clip}}  & \multicolumn{2}{c}{\textcolor{gray}{76.4}} & \multicolumn{2}{c}{\textcolor{gray}{91.5}} & \multicolumn{2}{c}{\textcolor{gray}{56.0}} & \multicolumn{2}{c}{\textcolor{gray}{73.6}} & \multicolumn{2}{c}{\textcolor{gray}{85.0}} & \multicolumn{2}{c}{\textcolor{gray}{89.6}} & \multicolumn{2}{c}{\textcolor{gray}{51.7}} & \multicolumn{2}{c}{\textcolor{gray}{73.6}} \\
\rowcolor{gray!10}\textcolor{gray}{SigLIP \cite{zhai2023siglip}} & \multicolumn{2}{c}{\textcolor{gray}{81.8}} & \multicolumn{2}{c}{\textcolor{gray}{90.5}} & \multicolumn{2}{c}{\textcolor{gray}{61.7}} & \multicolumn{2}{c}{\textcolor{gray}{76.1}} & \multicolumn{2}{c}{\textcolor{gray}{92.6}} & \multicolumn{2}{c}{\textcolor{gray}{95.1}} & \multicolumn{2}{c}{\textcolor{gray}{69.2}} & \multicolumn{2}{c}{\textcolor{gray}{89.1}} \\
\bottomrule
\end{tabular}
}
\vspace{-4pt}
\caption{OW results averaged on the grouped datasets. TI stands for text inclusion, LI for Llama inclusion, SS for semantic similarity, and CS for concept similarity. Higher is better, \textbf{bold} indicates best.}
\vspace{-12pt}
\label{tab:grouped_all}
\end{table*}

%% file: sec/4_analyses.tex
\label{sec:analyses}

\begin{figure*}
\centering
\resizebox{\linewidth}{!}{
\begin{tabular}{cccc}
{\includegraphics[trim={0
0 0.6cm 0}, clip, height=3.5cm]{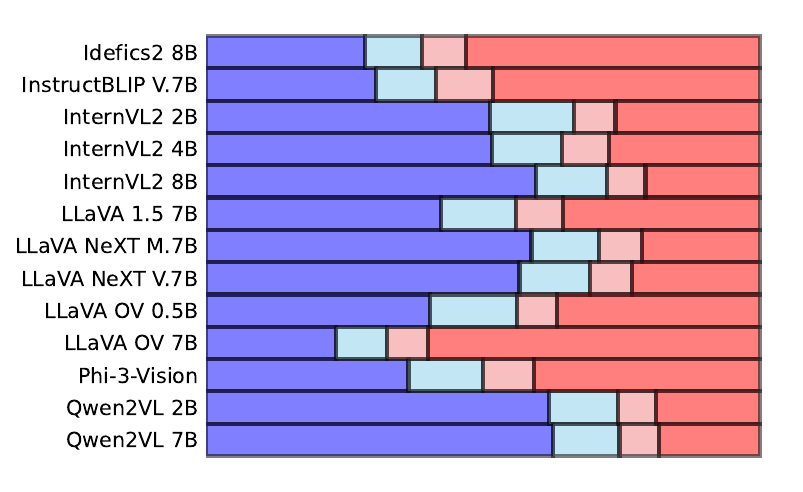}} &
{\includegraphics[trim={3.4cm
0 0.6cm 0}, clip, height=3.5cm]{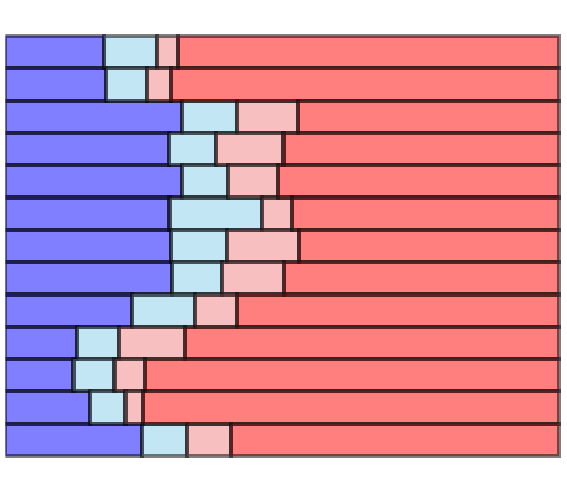}} &
{\includegraphics[trim={3.4cm
0 0.6cm 0}, clip, height=3.5cm]{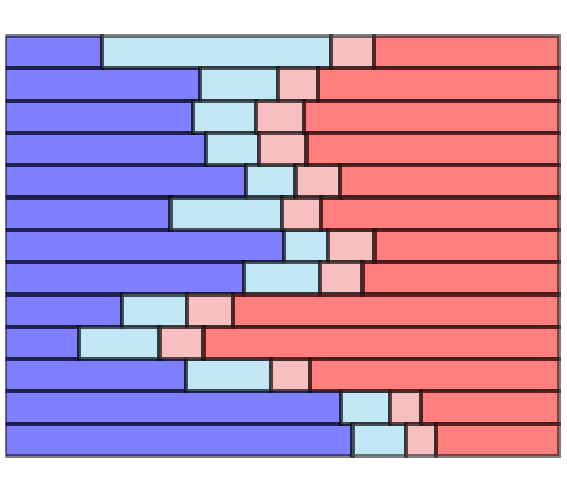}} &
{\includegraphics[trim={3.4cm
0 0.6cm 0}, clip, height=3.5cm]{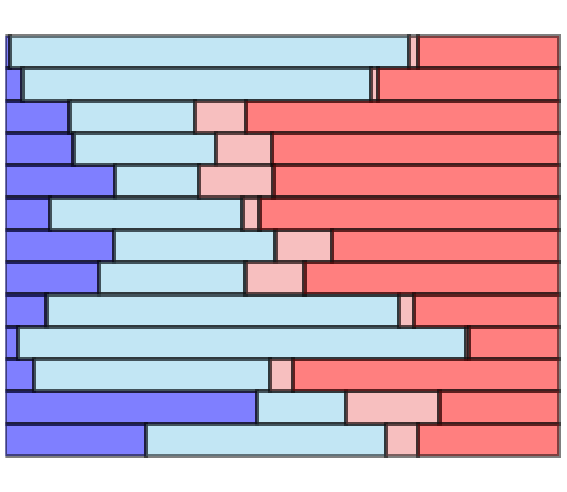}} \\
 (a) Prototypical & (b) Non-prototypical & (c) Fine-grained & (d) Very fine-grained \\
\end{tabular}}
\vspace{-8pt}
\caption{
    Types of model predictions per dataset groups.
    Blue indicates \inlineColorbox{blue!50}{correct and specific} and \inlineColorbox{skyblue!50}{correct but generic} predictions, red indicates \inlineColorbox{lightcoral!50}{wrong but specific} and \inlineColorbox{red!50}{wrong and generic} mistakes.
}
\vspace{-10pt}
\label{fig:group_errors_base}
\end{figure*}

\subsection{Interpreting Model Predictions Through Inclusion and Similarity Scores}
To underscore the importance of jointly evaluating inclusion and similarity scores, we present qualitative results demonstrating how their combined analysis offers deeper insights into LMM failures.
\cref{fig:metrics_per_image_qualitative} showcases qualitative results from various datasets, displaying ground truth class labels alongside model predictions.
For instance, the challenging case of \textit{caprese salad} illustrates this distinction: more descriptive predictions like \textit{creamy sauce}, which LI considers incorrect, receive a relatively higher CS score than \textit{food}, which is deemed correct by LI.
This emphasizes that \textit{creamy sauce} is semantically closer to the ground truth, yet it is rejected by LI due to its lack of alignment with the ground truth.
Similar behavior is present in the other examples.

To reinforce our previous point, Fig.~\ref{fig:metrics_per_class_qualitative} illustrates the relationship between LI and CS, highlighting the distinct contributions of these two metrics.
Predictions in the top-right quadrant correspond to concepts that are semantically close to the ground truth and are also likely to be considered correct by LI (\eg, \textit{cellphone} with predictions such as \textit{mobile phone} and \textit{Nokia}).
In contrast, the bottom-left quadrant represents the opposite case.
For instance, in the same plot, \textit{handheld device}—while somewhat related to \textit{cellphone} and receiving a nonzero CS score--is still deemed incorrect by LI.
Similarly, in the \textit{Caesar salad} example, the prediction \textit{food} appears in the bottom-right quadrant, as it is correct but overly generic.
Meanwhile, \textit{pasta salad}, being more specific yet incorrect, falls into the top-left quadrant.

\subsection{Grouping Model Predictions}

Following the intuition from above, we analyze the performance of LMMs defining four different groups of predictions: \corrspec, \corrgen~(\eg, \textit{dog} vs \textit{pug}); \wrongspec, predicting classes semantically similar to the target (\eg, \textit{pug} vs \textit{pomeranian}); and \wronggen~ \ie, where the prediction is semantically dissimilar from the target (\eg, \textit{sofa} vs \textit{dalmatian}).
To define these groups, we split the model predictions into four sets by thresholding the LI and CS scores.
We arbitrarily set the CS threshold at 0.6 to distinguish between generic and specific responses and the LI threshold at 0.5 to separate correct and wrong responses\footnote{Note that LI is either 0 or 1 on a per-sample basis, but it ranges between the two when considering aggregated results, \eg, average per class.}.
We visualize the ratios for the predictions in Fig.~\ref{fig:group_errors_base}.
Intuitively, a good LMM should have an high amount of predictions as \corrspec.
When not possible, however, having an equally high \corrgen~ratio is still better than having errors of any form.

In terms of optimal predictions, we see that the best-performing models vary according to dataset groups.
For prototypical classification, the models with the lowest error are InternVL~8B, Qwen2VL~2B, and Qwen2VL~7B.
For non-prototypical tasks, instead, LLaVA~1.5~7B performs best, but InternVL~2B and InternVL~8B provide slightly more precise predictions.
For fine-grained, trends are similar to the prototypical groups, but with fewer correct and more generic responses.
This is most evident for Idefics2~8B, which works fairly well on fine-grained classification but provides responses lacking specificity.
On very fine-grained, we perceive higher rates of \wronggen, with more generic predictions across all models.
Notably, Qwen2VL models perform better in the last two settings.
On average, the models with the highest wrong predictions are LLaVA-OV~7B and InstructBLIP Vicuna~7B.
The model that is, on average, more generic in its replies is Idefics2~8B.

\section{Analyzing LMMs Mistakes in OW }

In the following, we further inspect the correct and wrong predictions of different models.
Specifically, each section will analyze one of the four cases: \corrspec~ (\cref{sec:exp-correct-precise}), \corrgen~(\cref{sec:exp-correct-generic}), \wrongspec~ (\cref{sec:exp-wrong-precise}), and \wronggen~ (\cref{sec:exp-wrong-generic}).

\subsection{Correct and Specific}
\label{sec:exp-correct-precise}
While this section describes successful cases, from \cref{sec:results} we know that models perform differently.
Thus, here we investigate whether LMMs share similar success cases.

\input{tab/correct_dataset}

\begin{figure}
\includegraphics[width=7.5cm]{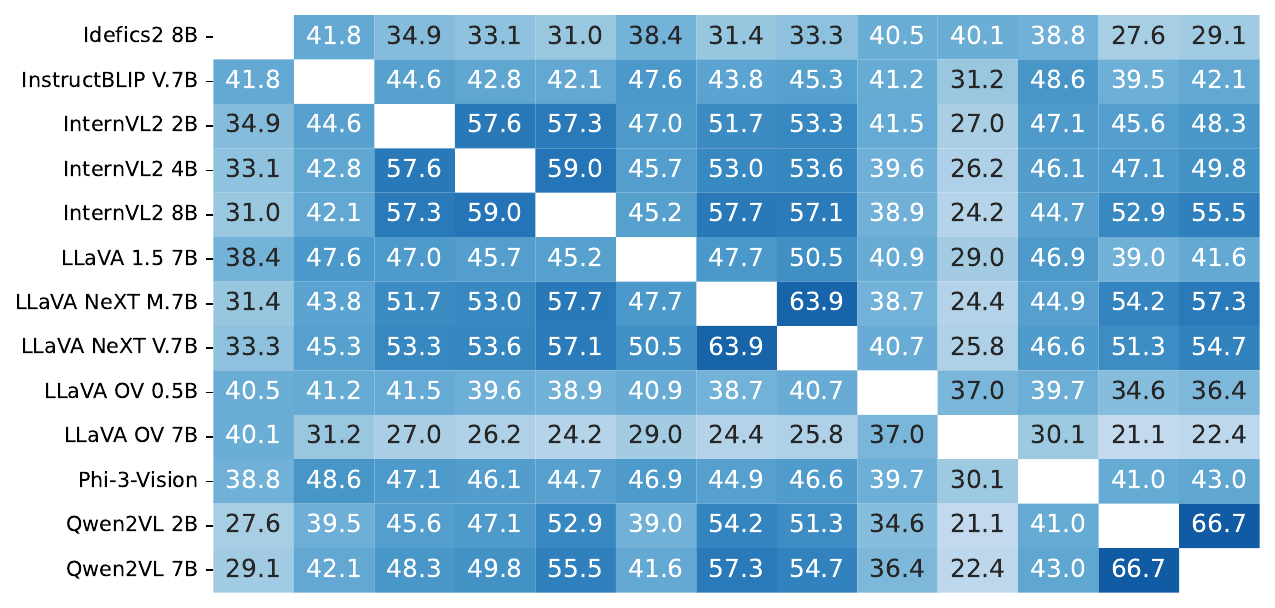}
\caption{
Percentage of correct and specific predictions shared between models.
Higher values indicate higher agreement.}

\vspace{-6pt}
\label{fig:conf_mat_common_correct_precise}
\vspace{-6pt}
\end{figure}

\begin{figure*}
\centering
\begin{tabular}{ccc}
{\includegraphics[trim={0
0 5 2.2cm}, clip, width=5.5cm]{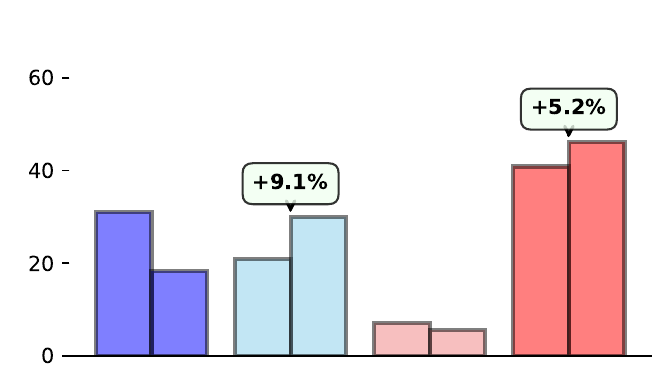}} &
{\includegraphics[trim={0
0 0 2.2cm}, clip, width=5.5cm]{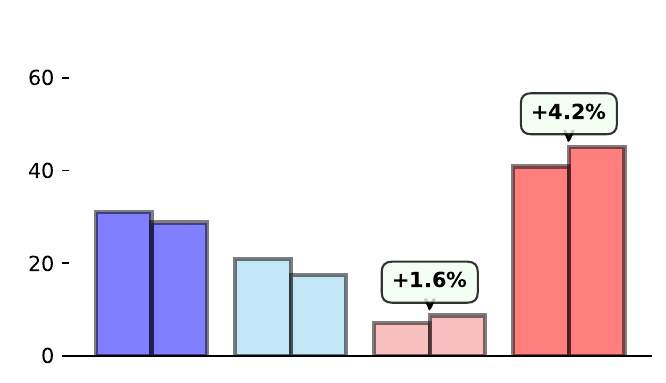}} &
{\includegraphics[trim={0
0 0 2.2cm}, clip, width=5.5cm]{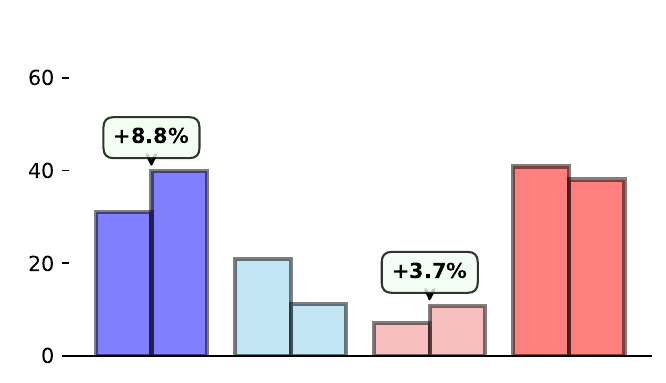}} \\
(a) Be generic & (b) Be specific & (c) Domain-specific \\
\end{tabular}
\vspace{-8pt}
\caption{
    Average gains per prediction type when asking models to be more generic/specific (a, b), or via dataset-specific prompts (c).
    Blue indicates \inlineColorbox{blue!50}{correct and specific} and \inlineColorbox{skyblue!50}{correct but generic} predictions, red indicates \inlineColorbox{lightcoral!50}{wrong but specific} and \inlineColorbox{red!50}{wrong and generic} mistakes.
}
\vspace{-10pt}
\label{fig:avg_group_error_gains}
\end{figure*}

\noindent\textbf{Are correct predictions shared among models?}
To answer this question, we first evaluate the percentage of samples that receive correct predictions by multiple models across datasets.
We report the results in \cref{tab:correct-datasets}, splitting them according to low (less than 30\% of models), medium (30\%-70\%), and high agreement (above 70\%).
The table shows that the models tend to agree on prototypical datasets (\eg, 71.4\% of high agreement on C101) but they do not for very fine-grained ones (\ie, CARS and FGVC).
Overall, we found that only 5.6\% of the samples are correctly predicted by all models and there exists 6 labels out of almost 1200 that are never predicted correctly according to the LI score: \ie, \textit{birman}, \textit{bishop of llandaff}, \textit{egyptian mau}, \textit{prince of wales feather}, \textit{silverbush}, and \textit{watercress}, all belonging to fine-grained datasets.
These results confirm the ability of LMMs to capture generic concepts while struggling on very specific ones.
In Fig.~\ref{fig:group_errors_base}, we observe that when the granularity constraint is relaxed, most models continue to predict the parent class with a remarkable level of accuracy, given the nature of the task.

\noindent\textbf{Which models agree the most with each other?} We additionally check the pair-wise agreement on the model predictions on the \corrspec~ group, showing the results in \cref{fig:conf_mat_common_correct_precise}.
Interestingly, models of the same family tend to share more predictions, \ie, Qwen2VL~2B and Qwen2VL~7B share 66.7\% \corrspec~ predictions, InternVL2 4B and InternVL2 8B 59.0\%.
This also happens with different language models (\eg, LLaVA NeXT with Mistral and Vicuna share 63.9\% of correct predictions), and differences might arise within lower performing families (\eg, LLaVA-OV~0.5B and 7B agree only 37\% of the time).
While there is no clear pattern, the best-performing families (\eg, LLaVA NeXT, InternVL2, Qwen2VL) tend to share more than half of the correct predictions (\eg, InternVL2 8B and Qwen2VL 7B, 55.5\%), suggesting that the agreement is mostly driven by the capabilities rather than design choices.
We also show the agreement between models on the other three prediction splits in the Supp. Mat. (see~\ref{sec:supp-analysis}).

\subsection{Correct but Generic}
\label{sec:exp-correct-generic}

As different classification scenarios may require different levels of granularity, in the following we check whether we can control the latter via prompting.
We investigate three types of requests: \textit{``Be generic.''}, \textit{``Be specific.''}, and \textit{domain}-specific prompts, focusing on the fine-grained and very fine-grained datasets, alongside DTD.
In \cref{fig:avg_group_error_gains}, we report the average difference across datasets for each group of predictions and type of prompt, reporting in the Supp. Mat. (see~\ref{sec:supp-analysis-extended}) the metric variations on datasets and models.

\noindent\textbf{Be more generic.} When queried for generic responses, we see a large shift from \corrspec~predictions to \corrgen, and, to a smaller degree, the same happens for wrong ones.
This highlights how models can provide good generic responses (+9.1\%) but the large decrease in \corrspec~ones means they become too generic.

\begin{figure*}
\centering
\begin{tabular}{cccc}
{\includegraphics[trim={0.3cm
0.3cm 0.3cm 2.2cm}, clip, width=4cm]{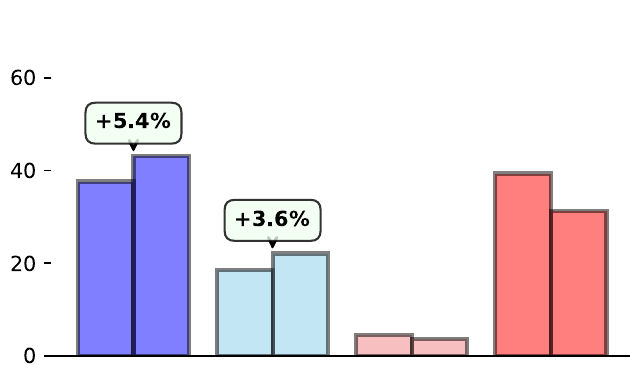}} &
{\includegraphics[trim={0.3cm
0.3cm 0.3cm 2.2cm}, clip, width=4cm]{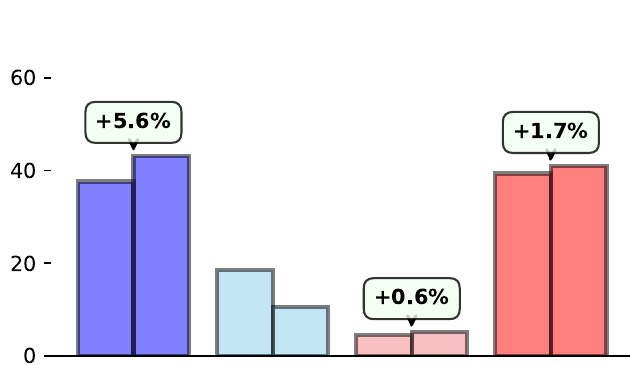}} &
{\includegraphics[trim={0.3cm
0.3cm 0.3cm  2.2cm}, clip, width=4cm]{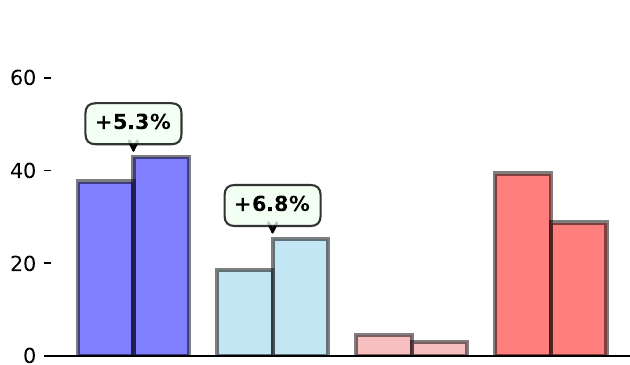}} & 
{\includegraphics[trim={0.3cm
0.3cm 0.3cm 2.2cm}, clip, width=4cm]{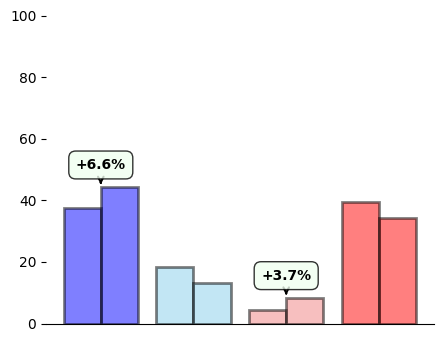}} \\
\\
(a) Zero-shot CoT & (b) LlamaV-o1 prompt & (c) LLaVA-CoT prompt & (d) Reasoning models \\
\end{tabular}
\vspace{-8pt}
\caption{
    Average gains per prediction types when including chain-of-thought reasoning (a, b, c), or when with reasoning models (d).
    Blue indicates \inlineColorbox{blue!50}{correct and specific} and \inlineColorbox{skyblue!50}{correct but generic} predictions, red indicates \inlineColorbox{lightcoral!50}{wrong but specific} and \inlineColorbox{red!50}{wrong and generic} mistakes.
}
\vspace{-10pt}
\label{fig:avg_group_error_gains_test_time}
\end{figure*}

\noindent\textbf{Be more specific.} In this case, all LMMs consistently get worse, equally increasing \wrongspec~and \wronggen~predictions.
While a decrease (especially in \corrgen) is expected, this hints that LMMs are stronger at providing more generic replies than more specific ones.

\noindent\textbf{Domain-specific.} When tackling specific fine-grained scenarios, it is possible to tailor a custom prompt, \eg, when classifying flowers, we can directly ask \textit{``What type of flower is in this image?''} instead of a generic object.
Therefore, we explore whether informing the LMM on the target fine-grained scenario may fix the specificity issue.
We update the prompt to use the terms ``texture'' (for DTD), ``aircraft'' (for FGVC), ``flower'' (for FLWR), ``food'' (for FOOD), ``pet'' (for PETS), or ``car'' (for CARS).
Overall, domain-specific prompts positively influence the predictions, converting an average of 12.5\% of generic responses into specific ones.
Notably, LLaVA-OV~0.5B gets +29\% on the \corrspec~set, followed by Qwen2VL~7B with +15\% (see~\ref{sec:supp-analysis-extended} in the Supp. Mat.).
This shows how, while LMMs struggle to provide specific predictions off-the-shelf, injecting domain-specific context can largely improve OW performance.

\subsection{Wrong but Specific}
\label{sec:exp-wrong-precise}

Here we analyze mistakes due to two objects being very similar (\eg, \textit{euphonium} vs \textit{trombone}).
As addressing this type of mistake requires reasoning on fine-level details of the images, we explore whether test-time reasoning can improve performance.
Thus, we study the impact of introducing Chain-of-Thought~\cite{wei2022cot,kojima2022zero-shot-cot} during inference.

\noindent\textbf{Can CoT mitigate misclassification?}
We identify three simple techniques we can apply without modifying the architecture of the models: zero-shot CoT~\cite{kojima2022zero-shot-cot} appending the instruction \textit{``Think step by step.''} to the input query, \textit{LlamaV-o1 prompt} using the multi-turn procedure of~\cite{thawakar2025llamav}, and the \textit{LLaVA-CoT prompt}~\cite{xu2024llavacot} for reasoning in procedures.
For this study, we focus on the InternVL2 and the Qwen2VL families, showing their average gains in \cref{fig:avg_group_error_gains_test_time}.
Additional results are available in the Supp. Mat. (see~\ref{sec:supp-analysis-extended}).
Notably, test-time reasoning helps the models in making \corrspec~responses.
Performance-wise, Qwen2VL shows the highest gains, achieving up to +13\% in \corrspec~responses.
While test-time reasoning enhance the OW of LMMs, we observe that the multi-turn prompt tends to steer the model either toward semantically correct predictions or completely divergent ones (+1.7 on \wronggen).
On the other hand, simply instructing LMMs to think with zero-shot CoT or providing a longer prompt (\textit{LLaVA-CoT}), consistently increases their accuracy.

\noindent\textbf{Do models tailored for reasoning excel in OW?} As we saw positive gains from using test-time reasoning, we further explore the capabilities of more advanced approaches.
Specifically, both InternVL2 and Qwen2-VL have two improved versions tailored for reasoning: InternVL2.5~\cite{chen2024internvl2.5} and Qwen2.5VL~\cite{bai2025qwen2.5}.
In the following, we check whether these variants outperform their predecessors, less tailored to reasoning.
We show the average relative gains in \cref{fig:avg_group_error_gains_test_time}~(d).

By directly replacing the base models with their reasoning counterparts, we get mixed results, as we see a large increase in correct prediction (+6.6\% on average), but also in misclassification with semantically close concepts (+3.7\%), the error we wanted to address.
This shows that test-time reasoning might be more effective at addressing such nuanced cases than reasoning-based models.

\begin{figure}
\includegraphics[trim={0
0.4cm 0 2.2cm}, clip, width=0.9\linewidth]{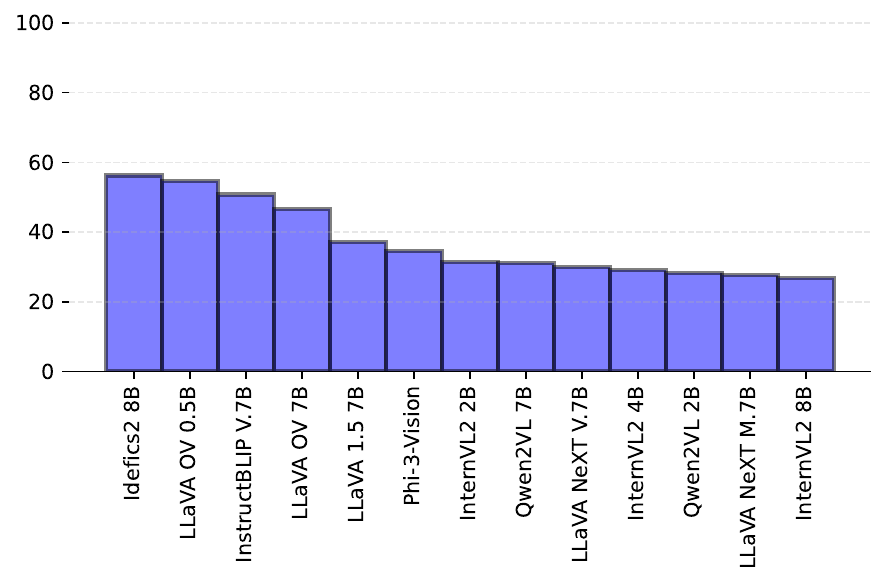}
\caption{
    Percentage of model predictions considered wrong in the single-label setting, but correct in multi-label.
}
\vspace{-10pt}
\label{fig:ram_matches}
\end{figure}

\subsection{Wrong and Generic}
\label{sec:exp-wrong-generic}
In this category,  predictions are not only wrong according to inclusion metrics but also based on semantic ones.
While some of the mistakes are due to the lack of fine-grained understanding of the models (see \cref{sec:exp-correct-precise}), here we investigate to which extent LMMs are correct even within wrong predictions.
Specifically, we explore cases where models simply focus on the wrong object in the image.

\noindent\textbf{Do LMMs focus on the wrong object?} To investigate this, we annotate images with multiple labels using RAM++~\cite{huang2023ramplus}, a state-of-the-art model for tagging images with a list of concepts.
Then, we compare LMM predictions to the list of tags, looking for cases where there is an extremely high CS (above 0.95 with any of the tags in the image).
If this is the case, we assume the prediction to be relevant for the image, even if different from the true label.

\cref{fig:ram_matches} shows the percentage of wrong predictions that match a tag.
As shown in the table, this percentage is high, ranging between 30\% and 60\% of the wrong predictions.
Notably, this is high also for models with lower overall performance in \cref{tab:grouped_all}, such as Idefics2 and InstructBLIP.
Additional experiments on the capability of models in predicting and suggesting multiple hypotheses for the output class are in Supp. Mat. (see~\ref{sec:supp-analysis}), where we explore their changes in accuracies when tasked to predict multiple labels.

%% file: tab/correct_dataset.tex
\begin{table}
  \centering
  \resizebox{!}{2cm}{
  \begin{tabular}{l|cccc}
    \toprule
    \multirow{2}{*}{\textbf{Dataset}} & \multicolumn{3}{c}{\textbf{Agreement}}\\
     & High (\%) & Medium (\%) & Low (\%) \\
    \midrule
    C101 & 71.4 & 15.8 & 12.8 \\
    S397 & 34.3 & 33.0 & 32.8 \\
    U101 & 33.8 & 26.8 & 39.5 \\
    FOOD & 32.6 & 27.5 & 39.9 \\
    {DTD} & 23.3 & 29.1 & 47.6 \\
    FLWR & 13.9 & 25.5 & 60.6 \\
    ESAT & 6.1 & 21.8 & 72.1 \\
    PETS & 5.5& 16.1 & 78.4  \\
    CARS & 1.5 & 21.9 &76.6 \\
    FGVC & 0.1 & 4.0 & 96.0  \\
   \bottomrule
  \end{tabular}}
  \caption{Agreement of LMMs correct predictions across datasets. Low indicates that less than 30\% of the models predicted a sample correctly, while high indicates that more than 70\% did.}
    \vspace{-5pt}
  \label{tab:correct-datasets}
\end{table}

%% file: sec/5_conclusions.tex
\section{Conclusions} 
\vspace{-2mm}
In this work, we conducted a large-scale study on LMMs for OW classification.
Evaluating 13 models across 10 datasets using four different metrics, we highlight both their strengths and the challenges they face in this task.
As the four metrics capture different levels of alignment between predictions and ground truth, we use them to provide an in-depth analysis of LMMs' mistakes, identifying cases where the model is too generic, confused by similar concepts, or focuses on the wrong subject, analyzing strategies to mitigate these issues.
Our benchmark and metrics can serve as a reference for future work in this field, toward tackling this challenging yet underexplored setting.

\section*{Acknowledgements}
This project was supported by PNRR ICSC National Research Centre for HPC, Big Data and Quantum Computing (CN00000013), FAIR - Future AI Research (PE00000013), funded by NextGeneration EU. This work is also supported by the EU projects ELIAS (101120237) and ELLIOT (101214398), and by Ministero delle Imprese e del Made in Italy (IPCEI Cloud DM 27 giugno 2022 – IPCEI-CL-0000007) and European Union (NextGeneration EU).

%% file: sec/6_supmat.tex
\appendix 
\section{Supplementary Material}
\label{sec:supmat}

In the following, we provide additional information on our analyses.
First, we report further detail on the considered datasets, models (\ref{sec:supp-dm}), and metrics (\ref{sec:supp-metrics}), followed by the extended results of each model for each dataset (\ref{sec:supp-results}).
Then, we extend our main analyses by evaluating with an Elo ranking system which model provides the best responses, using a Llama instance to score wins.
We continue the analysis by evaluating the percentage of agreement between models for the three prediction groups not present in the main paper, and we use RAM++ to tag images by checking whether we can improve the model performance by using prompts that foster multi-label responses, \eg, listing the objects in the scene, or describing the image (\ref{sec:supp-analysis}).
Finally, we report additional tables and visualizations to accompany the studies in the main manuscript (\ref{sec:supp-analysis-extended}).

\subsection{Additional details on the datasets and models}
\label{sec:supp-dm}
The datasets used in our evaluation are summarized in \cref{tab:datasets}.
For the experiments we used the same training and test splits used in previous works \cite{conti2023vocabulary}, while a summary of the LMMs used in this study and their differences is in \cref{tab:models_info}.

\subsection{Additional details on the metrics}
\label{sec:supp-metrics}
For computing the inclusion metric, we instruct Llama 3.2~\cite{grattafiori2024llama} to score good and bad LMM responses with the following prompt:

\begin{tcolorbox}[breakable, enhanced jigsaw, title=Llama inclusion instruction]
You are a model that determines whether an answer is a good reply to a question given also its target value. \\
\\
This is the question: What type of object is in this image? \\
This is the answer: \%s \\
This is the target value: \%s \\
\\
If the answer describes the target, reply positively. If the answer includes the target value or a synonym of it, reply positively. If the target is generic but it is related to the answer, reply positively. Reply only with "1" if yes, or "0" if no.
\end{tcolorbox}

\input{tab/datasets}
\input{tab/models_info}

\subsection{Extended results}
\label{sec:supp-results}
We report the per-dataset results of the evaluated LMMs, split into one table for each of the considered metrics, \ie, text inclusion in \cref{tab:textual_inclusion}, Llama inclusion in \cref{tab:llama_inclusion}, semantic similarity in \cref{tab:semantic_similarity}, and concept similarity in \cref{tab:concept_similarity}.

\input{tab/text_inclusion}
\input{tab/llama_inclusion}
\input{tab/semantic_similarity}
\input{tab/concept_similarity}

\subsection{Additional analyses}
\label{sec:supp-analysis}

\noindent\textbf{Which model provides the best responses?} To analyze which model provides the best responses, we compare their generations in pairs.
Specifically, for each of the ten datasets, we randomly sample 10'000 pairs of generations, and instruct a Llama 3.2 model to identify the best response in the pair, similarly to what done in the Chatbot Arena~\cite{chiang2024chatbot} but through automatic evaluation with LLM-as-a-judge~\cite{zheng2023judging}. We use the following prompt to instruct Llama 3.2 to judge the pairs of predictions and decide for a win:

\begin{tcolorbox}[breakable, enhanced jigsaw, title=Llama Elo ranking]
You are a model that discriminates whether labels A or B better align with a target value. \\
\\
This is label A: \%s \\
This is label B: \%s \\
This is the target value: \%s \\
\\
Does A align better with the target value? Does B align better with the target value? Reply only with "1" if A wins over B, or "0" if B wins over A.
\end{tcolorbox}

We directly compare the quality of the outputs by evaluating the Elo score~\cite{elo1967proposed} of these model responses and report the average on the ten datasets in \cref{tab:elo_ranking_average}.
Results show that Qwen2VL models are the best at providing accurate predictions, similar to the trend in \cref{tab:grouped_all}.

\input{tab/average_elo_ranking}

\begin{figure}
\includegraphics[width=8cm]{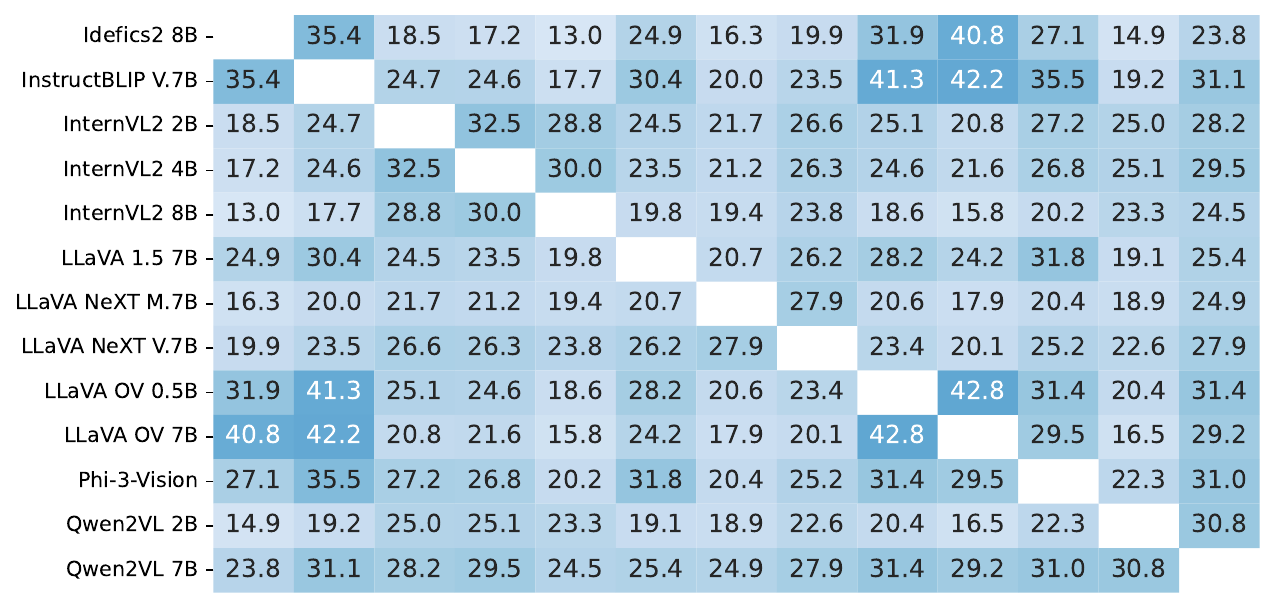}
\caption{
Percentage of correct but generic predictions shared between models.
Higher values indicate models perform responses similarly to the same inputs.}
\label{fig:conf_mat_common_correct_generic}
\end{figure}

\begin{figure}
\includegraphics[width=8cm]{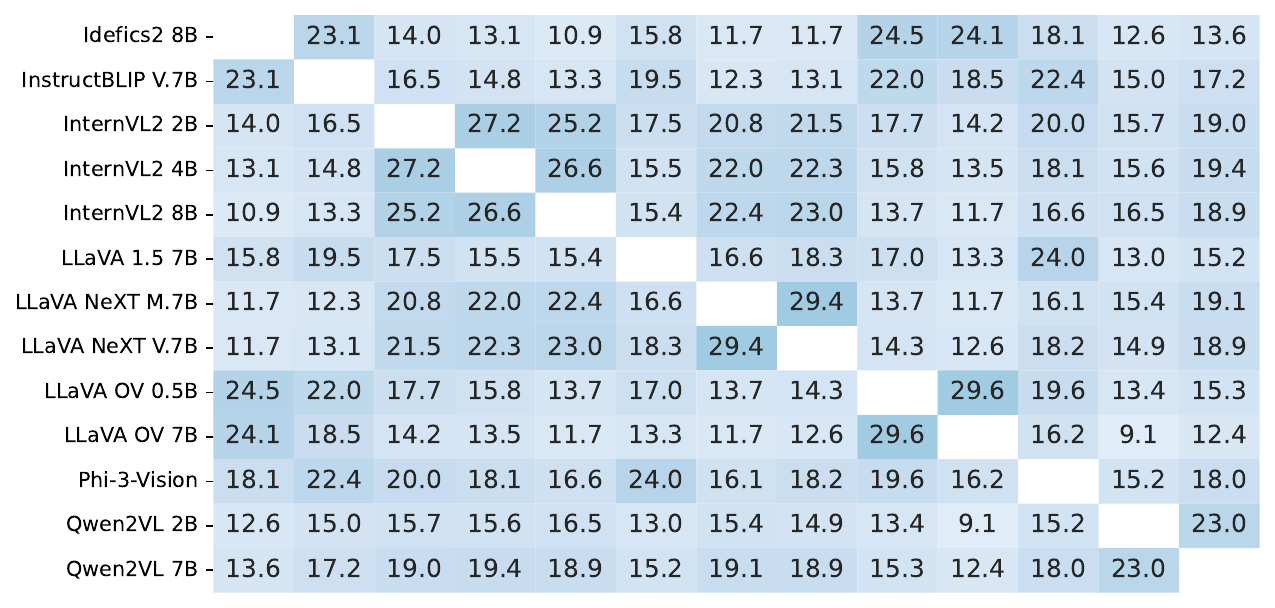}
\caption{
Percentage of wrong but specific predictions shared between models.
Higher values indicate models perform responses similarly to the same inputs.}
\label{fig:conf_mat_common_wrong_specific}
\end{figure}

\begin{figure}
\includegraphics[width=8cm]{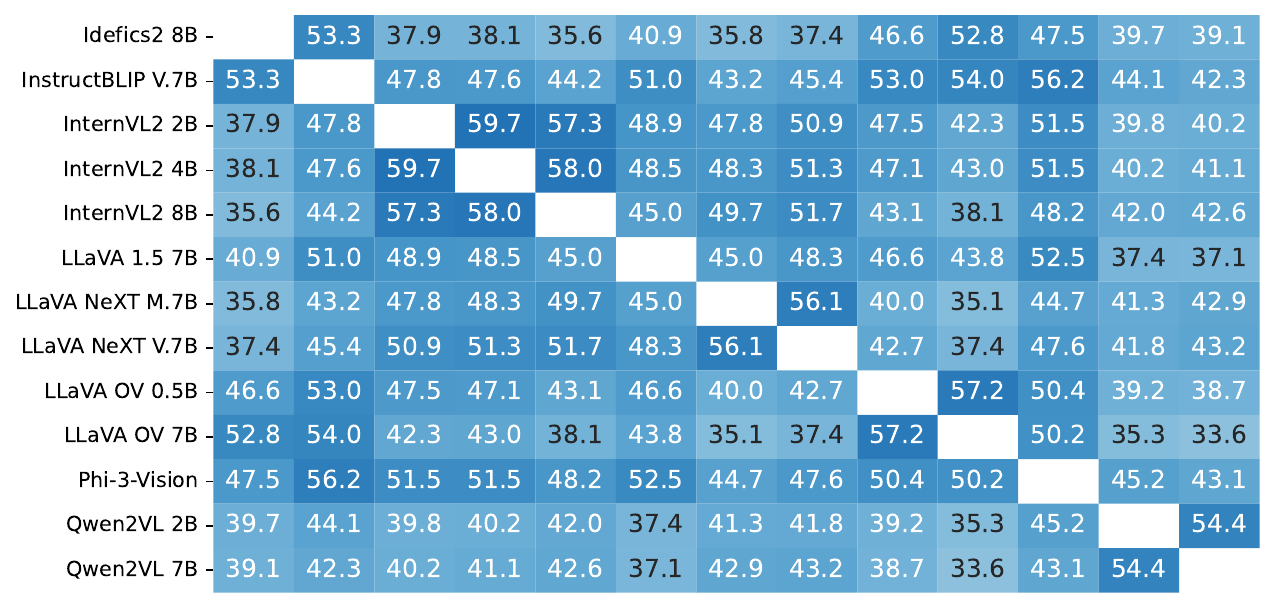}
\caption{
Percentage of wrong and generic predictions shared between models.
Higher values indicate models perform responses similarly to the same inputs.}
\label{fig:conf_mat_common_wrong_generic}
\end{figure}

\noindent\textbf{Which models agree the most with each other?} To complement the analysis of the main paper, here we show the pair-wise agreement on the model predictions on group beyond the correct and specific one, showing the results in \cref{fig:conf_mat_common_correct_generic} for correct but generic, \cref{fig:conf_mat_common_wrong_specific} for wrong but specific, and \cref{fig:conf_mat_common_wrong_generic} for wrong and generic.
The trends follow those of the main paper (\cref{fig:conf_mat_common_correct_precise}) \ie, where models of the same families tend to agree on the same samples, generalizing those findings across groups.

\noindent\textbf{Predicting more concepts.} The experiment using RAM++ to tag images suggested that LMMs often fail to predict the class names because they focus on the wrong part of the image.
However, when prompted to provide multiple candidates, do LMMs get the correct prediction?
To investigate this, we ask the model to (i) list the objects in the image; (ii) caption it, or (iii) describe its content.
We report the relative gain per model in \cref{tab:list_caption_describe}.
The results show that providing outputs that focus on multiple labels on average improves the concept-based similarity, with the only exception of the caption case.
Text inclusion improves consistently, showing that predictions become correct even according to this strict metric.
Overall, these results highlight how LMM mistakes can be ascribed by mismatches between the label and the focus of the annotator, with the models often focusing on grounded image content even in case of mistakes.

\input{tab/grouped_all_extended}
\input{tab/larger_and_commercial_models}

\noindent\textbf{Larger models.}
In \cref{tab:grouped_all_extended}, we add 6 larger models (\inlineColorbox{green!7}{green}) to the original 13 base, with scales from 13B to 72B.
Notably, \emph{scaling has mixed impacts}, sometimes leading to better performance (\eg, InternVL2 26B, Qwen2-VL 72B) and sometimes worse (\eg, InstructBLIP 13B, LLaVA-NeXT 34B).
Particularly, the language encoder in LLaVA-NeXT changes between 13B (Vicuna) and 34B (Yi), highlighting that \textit{the pre-training data has a stronger effect than scaling}.

\noindent\textbf{Commercial models.}
In \cref{tab:larger_and_commercial}, we report results for commercial models on a subset of the considered datasets. We compare these models against all the previously considered open-source models (\ie, 13 base + 5 reasoning and the larger models from the previous analysis).
The estimated sizes of these models are 8B (Haiku and GPT-4o-mini), 32B (Gemini 2.0 Flash), and +175B (Sonnet and GPT-4o). From the results, we notice there isn't a large gap between open and commercial models, with GPTs and Gemini performing on par with, \eg, InternVL2 26B and Qwen2-VL 72B. Only Claude consistently achieves better performance, still comparable to Qwen2.5-VL 7B at a fraction of the size. Surprisingly, GPT-4o-mini is better than GPT-4o on the task, similarly to the findings for InternVL2 2B \vs~4B. Also, \emph{reasoning models are strong}: Qwen2.5-VL 7B outperforms Qwen2-VL 72B and the commercial models on most of the metrics despite its reduced dimension.

\noindent\textbf{Linking model info with performance.}
Many models do not disclose their full training details, making it hard to identify key factors influencing performance. However, by linking the results in \cref{tab:grouped_all} with the summary in \cref{tab:models_info}, we hypothesize that (i) the pre-training of the vision encoder is more important than the size (\eg, InternVL and Qwen2-VL \vs~CLIP/SigLIP, or \vs~BLIP-2, which has double the size); (ii) higher image resolution can improve performance (\eg, LLaVA-NeXT \vs~LLaVA-1.5); (iii) the pre-training of the language encoder is less important than the training strategy (\eg, LLaVA-NeXT Mistral/Vicuna \vs~Idefics/InstructBLIP); (iv) the size of the language encoder is not an indicator of performance (\eg, LLaVA-NeXT Mistral \vs~Yi, GPT-4o-mini \vs~GPT-4o). It is also reasonable to assume that the strongest influence comes from the training data for which details are only partly available.

\subsection{Extended results for the analyses}
\label{sec:supp-analysis-extended}
\begin{figure*}
\centering
\resizebox{\linewidth}{!}{
\begin{tabular}{cccc}
{\includegraphics[trim={0
0 0.6cm 0}, clip, height=3.5cm]{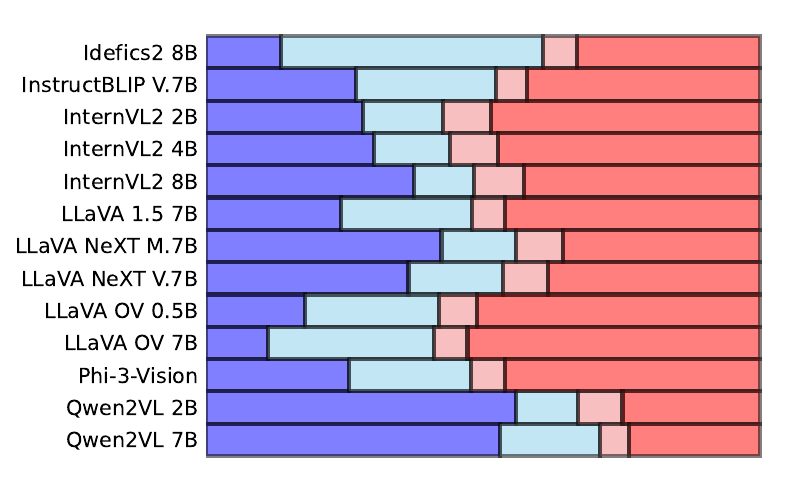}} &
{\includegraphics[trim={3.4cm
0 0.6cm 0}, clip, height=3.5cm]{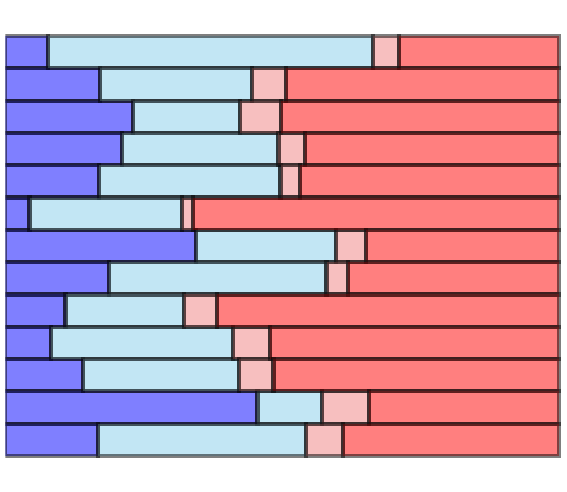}} &
{\includegraphics[trim={3.4cm
0 0.6cm 0}, clip, height=3.5cm]{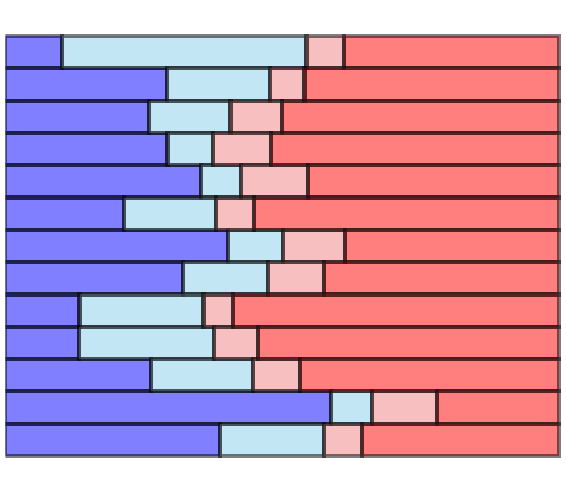}} &
{\includegraphics[trim={3.4cm
0 0.6cm 0}, clip, height=3.5cm]{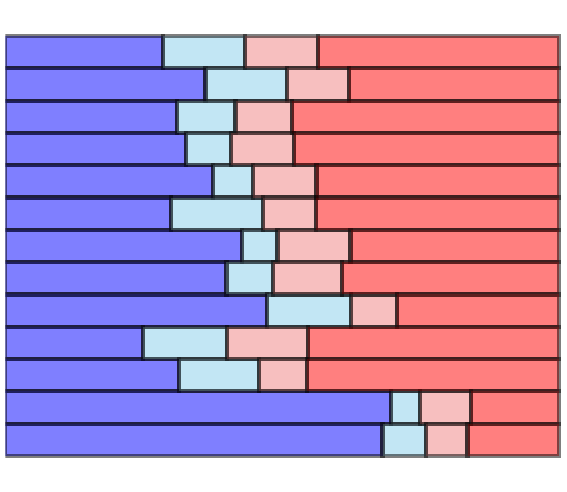}} \\
 (a) Default prompt & (b) Ask for generic & (c) Ask for specific & (d) Dataset prompts \\
\end{tabular}}
\caption{Types of model predictions when using the generic and specific prompts and the dataset-specific prompts.
Blue indicates \inlineColorbox{blue!50}{correct and specific} and \inlineColorbox{skyblue!50}{correct but generic} predictions, red indicates \inlineColorbox{lightcoral!50}{wrong but specific} and \inlineColorbox{red!50}{wrong and generic} mistakes.}
\label{fig:group_errors_zs_generic_specific_domain}
\end{figure*}

\begin{figure*}
\centering
\resizebox{\linewidth}{!}{
\begin{tabular}{cccc}
{\includegraphics[trim={0
0 0.6cm 0}, clip, height=3.5cm]{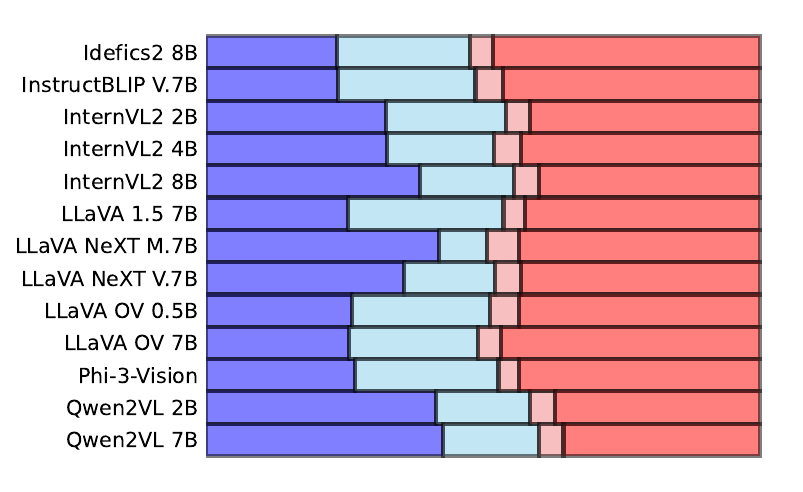}} &
{\includegraphics[trim={3.4cm
0 0.6cm 0}, clip, height=3.5cm]{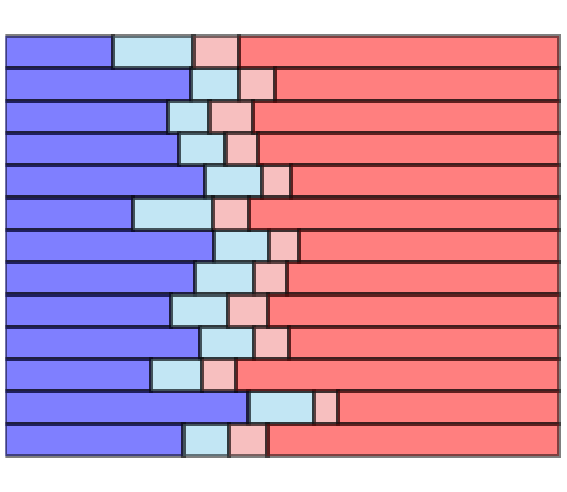}} &
{\includegraphics[trim={3.4cm
0 0.6cm 0}, clip, height=3.5cm]{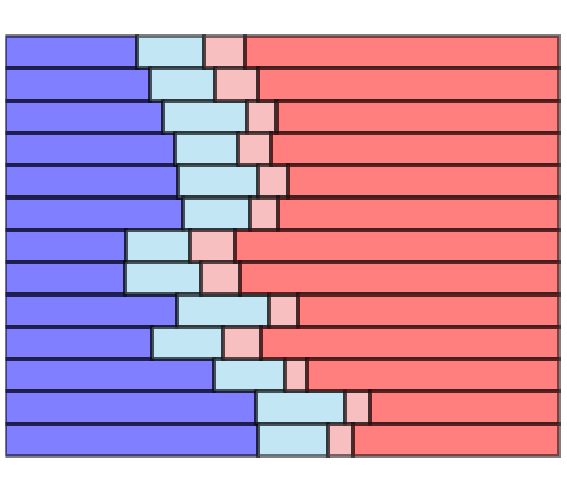}} &
{\includegraphics[trim={3.4cm
0 0.6cm 0}, clip, height=3.5cm]{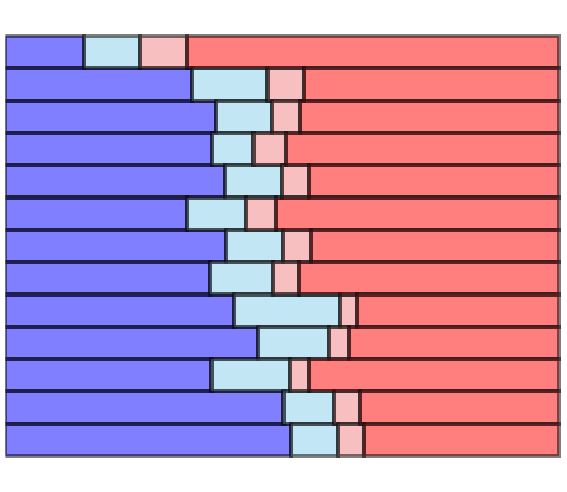}} \\
 (a) Default prompt & (b) List objects & (c) Caption & (d) Describe \\
\end{tabular}}
\caption{Types of model predictions when using multi-label prompts.
Blue indicates \inlineColorbox{blue!50}{correct and specific} and \inlineColorbox{skyblue!50}{correct but generic} predictions, red indicates \inlineColorbox{lightcoral!50}{wrong but specific} and \inlineColorbox{red!50}{wrong and generic} mistakes.}
\label{fig:group_errors_list_caption_describe}
\end{figure*}

Below, we report the extended results for the analyses we conducted.
In \cref{tab:group_errors_more_generic} (also visualizing the average gains in \cref{fig:group_errors_zs_generic_specific_domain}) we show the variation in correct and wrong predictions for each model when using more generic/specific prompts and domain-specific information.
We additionally report the variation in text inclusion, Llama inclusion, and concept similarity for each model and dataset in \cref{tab:more_generic_more_specific_relative} and \cref{tab:fine_grained_relative}.
For the chain-of-thought experiments, we provide the variations on the correct and wrong predictions in \cref{tab:group_errors_test_time}, and the per-dataset and model variations in \cref{tab:test_time}.
We also provide a visualization of the variations for the list, caption, and describe experiments in \cref{fig:group_errors_list_caption_describe} (also reported numerically in \cref{tab:group_errors_list_caption_describe}).
Finally, we report the complete results table for the reasoning models tested on the ten classification datasets in \cref{tab:reasoning_models}.

\include{tab/list_caption_describe}

\input{tab/generic_specific}
\input{tab/fine_grained}
\input{tab/group_errors_generic_specific}

\input{tab/test_time}

\input{tab/group_errors_chain_of_thought}

\include{tab/group_errors_list_caption_describe}

\input{tab/reasoning_models}

%% file: tab/datasets.tex
\begin{table}
  \centering
  \begin{tabular}{l|cc}
    \toprule
    \textbf{Dataset} & \textbf{Images} & \textbf{Classes} \\
    \midrule
    \textsc{Caltech101} \cite{fei2004caltech101}  (C101) & 2,465 & 100 \\
    \textsc{DTD} \cite{cimpoi2014dtd} & 1,692 & 47 \\
    \textsc{EuroSAT} \cite{helber2019eurosat} (ESAT) & 8,100 & 10 \\
    \textsc{FGVCAircraft} \cite{maji2013aircraft} (FGVC) & 3,333 & 100 \\
    \textsc{Flowers102} \cite{nilsback2008flowers} (FLWR) & 2,463 & 102 \\
    \textsc{Food101} \cite{bossard2014food} (FOOD) & 30,300 & 101 \\
    \textsc{OxfordPets} \cite{parkhi2012pets} (PETS) & 3,669 & 37 \\
    \textsc{Stanford Cars} \cite{krause20133cars} (CARS) & 8,041 & 196 \\
    \textsc{SUN397} \cite{xiao2010sun} (S397) & 19,850 & 397 \\
    \textsc{UCF101} \cite{soomro2012ucf101} (U101) & 3,783 & 101 \\
   \bottomrule
  \end{tabular}
  \caption{Summary details of the datasets used in our analyses.}
  \label{tab:datasets}
\end{table}

%% file: tab/models_info.tex
\begin{table*}
\centering
\resizebox{\linewidth}{!}{
\begin{tabular}{p{2.9cm}|p{3cm}|p{2.6cm}|p{4cm}|p{4cm}}
    \toprule
    \textbf{Model} & \textbf{Vision Enc} & \textbf{Language Enc} & \textbf{Training} & \textbf{Pre-training} \\
    \midrule
    \textsc{Idefics2} \cite{laurenccon2025idefics} & \small SOViT (SigLIP), 0.4B params; max 980x980. & \small Mistral 7B & \small Interleaved web docs, image-caption pairs (LAION-COCO), OCR data; fine-tuned on 50 curated datasets. & \small Joint dual encoder training with Perceiver pooling for vision-text alignment. \\
    \midrule
    \textsc{InstructBLIP} \cite{dai2023instructblip} & \small ViT-g (BLIP-2), 1.1B params; 224x224. & \small Vicuna 7B & \small 26 datasets transformed into instruction-tuning format: captioning, VQA, image generation. & \small Two-stage pre-training: Vision-language alignment via BLIP-2 and instruction-aware Query Transformer for task-specific feature extraction. \\
    \midrule
    \textsc{InternVL2} \cite{chen2024internvl} & \small InternViT (custom), 0.3B params (or 6B for larger models); dynamic resolution, max 40 tiles of 448×448. & \small Qwen2 0.5B (for 1B and 2B versions), or InternLM2 8B (for 8B version). & \small Interleaved image-text, multilingual OCR, mathematical charts; strict quality control. & \small Progressive training: masked video modeling, cross-modal contrastive learning, and next-token prediction with spatiotemporal focus. \\
    \midrule
    \textsc{LLaVA-1.5} \cite{liu2023llava} & \small ViT-L (CLIP), 0.3B params; 336x336. & \small Vicuna 7B &\small 158K multimodal instruction-following samples; pre-trained on filtered CC dataset (596K image-text pairs). & \small Frozen vision encoder during feature alignment stage; end-to-end fine-tuning. \\
    \midrule
    \textsc{LLaVA-NeXT} \cite{li2024llavaNext} & \small ViT-L (CLIP), 0.3B params; 336x336, 672x672, 336x1344, and 1344x336. & \small Mistral 7B, or Vicuna 7B & \small Diverse tasks, including multi-image and video understanding. & \small Builds on LLaVA with extended ViT and additional multimodal datasets for improved generalization. \\
    \midrule
    \textsc{LLaVA-OV} \cite{li2024llava-ov} & \small SOViT (SigLIP), 0.4B params; dynamic resolution (AnyRes-9), max 2304x2304. & \small Qwen2 0.5B, or Qwen2 7B & \small Single-image and video scenarios with task transfer capabilities; diverse visual benchmarks. & \small Pre-trained with balanced visual token representation across scenarios to enable task transfer. \\
    \midrule
    \textsc{Phi-3-Vision} \cite{abdin2024phi} & \small ViT-L (CLIP), 0.4B params; dynamic resolution, max 1344x1344. & \small Phi-3 Mini (3.8B params) & \small Synthetic data, filtered public docs, high-quality interleaved text-image data, math/code examples. & \small Multi-stage training: custom vision encoder aligned with Phi-3 Mini language model using interleaved and fine-grained tasks. \\
    \midrule
    \textsc{Qwen2VL} \cite{wang2024qwen2} & \small ViT (custom), 0.6B params; dynamic resolution (Naive Dynamic Resolution), no max. & \small Qwen2 1.5B, or Qwen2 7B & \small Multilingual datasets: MathVista, DocVQA, RealWorldQA; supports videos (20+ min) and multilingual text in images. & \small Pre-trained with dynamic resolution ViT for flexible input sizes and multilingual alignment strategies. \\
    \bottomrule
\end{tabular}}
\caption{Summary details of the Language Multimodal Models used in our analyses.}
\label{tab:models_info}
\end{table*}

%% file: tab/text_inclusion.tex
\begin{table*}
  \centering
  \resizebox{\linewidth}{!}{
  \begin{tabular}{l|cccccccccc|c}
    \toprule
    & \multicolumn{11}{c}{\textbf{Textual inclusion}} \\
    \textbf{Model} & C101 & DTD & ESAT & FGVC & FLWR & FOOD & PETS & CARS & S397 & U101 & Avg. \\
    \midrule
    \rowcolor{cyan!7} \textsc{Idefics2} \cite{laurenccon2025idefics} 8B & 52.0 & 1.7 & 1.6 & 0.0 & 0.8 & 8.2 & 0.1 & 0.0 & 9.6 & 7.9 & 8.2 \\
    \textsc{InstructBLIP} \cite{dai2023instructblip} Vicuna 7B & 47.8 & 3.0 & 5.5 & 0.0 & 6.0 & 24.3 & 0.8 & 0.0 & 11.6 & 9.6 & 10.9 \\
    \rowcolor{cyan!7} \textsc{InternVL2} \cite{chen2024internvl, chen2024far} 2B & 52.8 & 10.8 & 7.4 & 1.4 & 14.1 & 23.3 & 7.2 & 0.0 & 21.1 & 12.4 & 15.0 \\
    \rowcolor{cyan!7} \textsc{InternVL2} \cite{chen2024internvl,chen2024far} 4B & 49.6 & 11.8 & 6.0 & 3.4 & 12.8 & 28.2 & 7.8 & 0.0 & 23.0 & 12.7 & 15.5 \\
    \rowcolor{cyan!7} \textsc{InternVL2} \cite{chen2024internvl,chen2024far} 8B & 55.0 & 12.5 & 6.0 & 4.6 & 19.1 & 33.9 & 13.8 & \textbf{0.1} & 26.3 & \textbf{14.4} & 18.6 \\
    \textsc{LLaVA-1.5} \cite{liu2023llava} 7B & 51.6 & 6.0 & \textbf{11.7} & 0.1 & 6.7 & 17.6 & 1.1 & 0.0 & 17.6 & 8.2 & 12.1 \\
    \textsc{LLaVA-NeXT} \cite{li2024llavaNext} (Mistral 7B) & 58.0 & 13.6 & 7.4 & 2.8 & 17.6 & 35.5 & 27.1 & 0.0 & 25.4 & 13.0 & 20.0 \\
    \textsc{LLaVA-NeXT} \cite{li2024llavaNext} (Vicuna 7B) & 54.9 & 12.2 & 7.2 & 2.5 & 11.9 & 29.6 & 9.4 & 0.0 & 24.0 & 12.5 & 16.4 \\
    \rowcolor{cyan!7} \textsc{LLaVa-OV} \cite{li2024llava-ov} (Qwen2 0.5B) & 53.4 & 9.2 & 4.2 & 1.2 & 2.9 & 12.6 & 2.5 & \textbf{0.1} & 15.5 & 8.7 & 11.0 \\
    \rowcolor{cyan!7} \textsc{LLaVa-OV} \cite{li2024llava-ov} (Qwen2 7B) & 55.5 & 12.6 & 4.9 & 0.0 & 14.2 & 5.0 & 0.1 & 0.0 & 6.2 & 4.0 & 10.2 \\
    \textsc{Phi-3-Vision} \cite{abdin2024phi} & 53.4 & 10.9 & 0.8 & 0.4 & 12.0 & 21.6 & 6.5 & \textbf{0.1} & 14.7 & 6.5 & 12.7 \\
    \rowcolor{cyan!7} \textsc{Qwen2VL} \cite{wang2024qwen2} 2B & 60.8 & 12.1 & 0.4 & \textbf{25.6} & \textbf{42.9} & 48.5 & 15.7 & \textbf{0.1} & 29.0 & 10.8 & \textbf{24.6} \\
    \rowcolor{cyan!7} \textsc{Qwen2VL} \cite{wang2024qwen2} 7B & \textbf{63.2} & \textbf{15.7} & 2.7 & 1.4 & 42.3 & \textbf{49.3} & 12.1 & \textbf{0.1} & \textbf{29.5} & 12.5 & 22.9 \\
    \midrule
    \rowcolor{pink!25} \multicolumn{12}{l}{\footnotesize \textit{Open-world baselines}} \\
    \rowcolor{pink!25} CaSED~\cite{conti2023vocabulary} & 35.5 & 5.1 & 3.0 & 1.4 & 28.1 & 19.4 & \textbf{34.6} & 0.0 & 13.5 & 8.1 & 14.9 \\
    \rowcolor{pink!25} CLIP retrieval & 42.6 & 7.5 & 6.6 & 14.0 & 40.6 & 26.4 & 30.3 & 0.0 & 14.7 & 8.4 & 19.1 \\
    \midrule
    \rowcolor{gray!10}\multicolumn{12}{l}{\footnotesize \textcolor{gray}{\textit{Closed-world baselines}}} \\
    \rowcolor{gray!10}\textcolor{gray}{CLIP~\cite{radford2021clip}} & \textcolor{gray}{87.1} & \textcolor{gray}{52.6} & \textcolor{gray}{42.7} & \textcolor{gray}{27.2} & \textcolor{gray}{76.9} & \textcolor{gray}{89.9} & \textcolor{gray}{88.1} & \textcolor{gray}{76.2} & \textcolor{gray}{65.6} & \textcolor{gray}{72.7} & \textcolor{gray}{67.9} \\
    \rowcolor{gray!10}\textcolor{gray}{SigLIP~\cite{zhai2023siglip}} & \textcolor{gray}{93.6} & \textcolor{gray}{60.8} & \textcolor{gray}{42.1} & \textcolor{gray}{46.0} & \textcolor{gray}{88.2} & \textcolor{gray}{94.1} & \textcolor{gray}{95.4} & \textcolor{gray}{92.3} & \textcolor{gray}{69.9} & \textcolor{gray}{82.1} & \textcolor{gray}{76.5} \\
    \bottomrule
  \end{tabular}}
  \caption{Text inclusion on the ten datasets. Higher is better, \textbf{bold} indicates best.}
  \label{tab:textual_inclusion}
\end{table*}

%% file: tab/llama_inclusion.tex
\begin{table*}
  \centering
  \resizebox{\linewidth}{!}{
  \begin{tabular}{l|cccccccccc|c}
    \toprule
    & \multicolumn{11}{c}{\textbf{Llama inclusion}} \\
    \textbf{Model} & C101 & DTD & ESAT & FGVC & FLWR & FOOD & PETS & CARS & S397 & U101 & Avg. \\
    \midrule
    \rowcolor{cyan!7} \textsc{Idefics2} \cite{laurenccon2025idefics} 8B & 72.9 & 24.6 & 19.0 & 64.4 & 54.6 & 58.7 & 36.3 & 69.6 & 32.5 & 40.1 & 47.3 \\
    \textsc{InstructBLIP} \cite{dai2023instructblip} Vicuna 7B & 76.8 & 26.2 & 19.1 & 59.9 & 57.4 & 47.6 & 41.3 & 62.0 & 35.8 & 36.0 & 46.2 \\
    \rowcolor{cyan!7} \textsc{InternVL2} \cite{chen2024internvl, chen2024far} 2B & 74.9 & 48.5 & 35.0 & 35.8 & 49.3 & 44.3 & 47.4 & 30.0 & 64.9 & 52.1 & 48.2 \\
    \rowcolor{cyan!7} \textsc{InternVL2} \cite{chen2024internvl,chen2024far} 4B & 74.4 & 45.7 & 30.1 & 40.5 & 37.5 & 45.9 & 49.7 & 33.1 & 62.5 & 50.4 & 47.0 \\
    \rowcolor{cyan!7} \textsc{InternVL2} \cite{chen2024internvl,chen2024far} 8B & 77.2 & 50.5 & 28.6 & 29.7 & 36.0 & 53.7 & 50.4 & 35.3 & 71.5 & \textbf{59.6} & 49.3 \\
    \textsc{LLaVA-1.5} \cite{liu2023llava} 7B & 74.5 & 39.4 & \textbf{45.0} & 44.5 & 46.3 & 47.7 & 45.5 & 37.5 & 51.6 & 48.5 & 48.1 \\
    \textsc{LLaVA-NeXT} \cite{li2024llavaNext} (Mistral 7B) & 77.8 & 54.0 & 28.0 & 43.4 & 33.4 & 63.2 & 34.6 & 50.9 & 69.9 & 58.3 & 51.4 \\
    \textsc{LLaVA-NeXT} \cite{li2024llavaNext} (Vicuna 7B) & 77.3 & 52.2 & 26.4 & 43.1 & 29.2 & 60.6 & 43.6 & 41.2 & 68.2 & 59.1 & 50.1 \\
    \rowcolor{cyan!7} \textsc{LLaVa-OV} \cite{li2024llava-ov} (Qwen2 0.5B) & 76.5 & 46.5 & 28.7 & 61.2 & 55.1 & 28.1 & 44.9 & 70.0 & 52.2 & 35.8 & 49.9 \\
    \rowcolor{cyan!7} \textsc{LLaVa-OV} \cite{li2024llava-ov} (Qwen2 7B) & 81.3 & 45.6 & 11.8 & \textbf{68.9} & 48.9 & 22.0 & 50.2 & \textbf{84.4} & 25.0 & 27.0 & 46.5 \\
    \textsc{Phi-3-Vision} \cite{abdin2024phi} & 75.7 & 45.3 & 6.0 & 51.0 & 53.2 & 45.1 & 49.1 & 39.0 & 44.5 & 34.7 & 44.4 \\
    \rowcolor{cyan!7} \textsc{Qwen2VL} \cite{wang2024qwen2} 2B & 82.9 & 54.6 & 3.1 & 65.0 & 67.0 & 71.1 & 49.3 & 56.3 & 72.6 & 45.2 & 56.7 \\
    \rowcolor{cyan!7} \textsc{Qwen2VL} \cite{wang2024qwen2} 7B & \textbf{84.3} & \textbf{60.8} & 18.1 & 58.8 & \textbf{71.0} & \textbf{75.0} & 46.0 & 67.2 & \textbf{73.0} & 48.8 & \textbf{60.3} \\
    \midrule
    \rowcolor{pink!25} \multicolumn{12}{l}{\footnotesize \textit{Open-world baselines}} \\
    \rowcolor{pink!25} CaSED~\cite{conti2023vocabulary} & 57.7 & 16.7 & 7.3 & 30.7 & 46.0 & 35.1 & \textbf{58.7} & 63.5 & 34.9 & 31.7 & 38.2 \\
    \rowcolor{pink!25} CLIP retrieval & 55.3 & 28.2 & 12.7 & 25.8 & 44.6 & 35.4 & 56.2 & 10.4 & 30.5 & 32.9 & 33.2 \\
    \midrule
    \rowcolor{gray!10}\multicolumn{12}{l}{\footnotesize \textcolor{gray}{\textit{Closed-world baselines}}} \\
    \rowcolor{gray!10}\textcolor{gray}{CLIP~\cite{radford2021clip}} & \textcolor{gray}{87.1} & \textcolor{gray}{52.6} & \textcolor{gray}{42.7} & \textcolor{gray}{27.2} & \textcolor{gray}{76.9} & \textcolor{gray}{89.9} & \textcolor{gray}{88.1} & \textcolor{gray}{76.2} & \textcolor{gray}{65.6} & \textcolor{gray}{72.7} & \textcolor{gray}{67.9} \\
    \rowcolor{gray!10}\textcolor{gray}{SigLIP~\cite{zhai2023siglip}} & \textcolor{gray}{93.6} & \textcolor{gray}{60.8} & \textcolor{gray}{42.1} & \textcolor{gray}{46.0} & \textcolor{gray}{88.2} & \textcolor{gray}{94.1} & \textcolor{gray}{95.4} & \textcolor{gray}{92.3} & \textcolor{gray}{69.9} & \textcolor{gray}{82.1} & \textcolor{gray}{76.5} \\
    \bottomrule
  \end{tabular}}
  \caption{Llama inclusion on the ten datasets. Higher is better, \textbf{bold} indicates best. Note that the scores for CLIP closed-world equals the textual inclusion scores.
}
  \label{tab:llama_inclusion}
\end{table*}

%% file: tab/semantic_similarity.tex
\begin{table*}
  \centering
  \resizebox{\linewidth}{!}{
  \begin{tabular}{l|cccccccccc|c}
    \toprule
    & \multicolumn{11}{c}{\textbf{Semantic similarity}} \\
    \textbf{Model} & C101 & DTD & ESAT & FGVC & FLWR & FOOD & PETS & CARS & S397 & U101 & Avg. \\
    \midrule
    \rowcolor{cyan!7} \textsc{Idefics2} \cite{laurenccon2025idefics} 8B & 64.9 & 34.6 & 27.5 & 27.6 & 38.6 & 44.4 & 30.8 & 31.6 & 44.2 & 44.0 & 38.8 \\
    \textsc{InstructBLIP} \cite{dai2023instructblip} Vicuna 7B & \textbf{71.5} & 32.8 & 30.0 & 21.4 & 38.9 & 41.6 & 26.4 & 38.5 & 42.1 & 48.3 & 39.1 \\
    \rowcolor{cyan!7} \textsc{InternVL2} \cite{chen2024internvl, chen2024far} 2B & 50.5 & 25.6 & 26.0 & 23.4 & 31.2 & 39.6 & 23.9 & 42.9 & 43.3 & 43.1 & 34.9 \\
    \rowcolor{cyan!7} \textsc{InternVL2} \cite{chen2024internvl,chen2024far} 4B & 49.2 & 26.1 & 24.7 & 23.6 & 30.2 & 41.1 & 24.6 & 44.1 & 43.8 & 41.8 & 34.9 \\
    \rowcolor{cyan!7} \textsc{InternVL2} \cite{chen2024internvl,chen2024far} 8B & 50.1 & 26.7 & 24.4 & 25.5 & 32.8 & 44.2 & 27.3 & 46.6 & 46.3 & 44.6 & 36.8 \\
    \textsc{LLaVA-1.5} \cite{liu2023llava} 7B & 49.0 & 24.2 & \textbf{34.2} & 19.0 & 25.8 & 37.2 & 21.5 & 38.2 & 41.7 & 40.7 & 33.1 \\
    \textsc{LLaVA-NeXT} \cite{li2024llavaNext} (Mistral 7B) & 48.2 & 27.7 & 23.9 & 23.6 & 30.2 & 45.3 & 30.3 & 44.8 & 43.6 & 42.1 & 36.0 \\
    \textsc{LLaVA-NeXT} \cite{li2024llavaNext} (Vicuna 7B) & 49.2 & 27.9 & 23.1 & 23.4 & 29.3 & 43.0 & 24.4 & 45.7 & 43.3 & 42.3 & 35.1 \\
    \rowcolor{cyan!7} \textsc{LLaVa-OV} \cite{li2024llava-ov} (Qwen2 0.5B) & 64.7 & 28.8 & 21.6 & 21.0 & 41.4 & 42.7 & 31.4 & 40.0 & 43.2 & 47.9 & 38.3 \\
    \rowcolor{cyan!7} \textsc{LLaVa-OV} \cite{li2024llava-ov} (Qwen2 7B) & 68.7 & 32.2 & 19.4 & 29.4 & 37.5 & 41.7 & 37.8 & 34.4 & 43.4 & 43.2 & 38.8 \\
    \textsc{Phi-3-Vision} \cite{abdin2024phi} & 53.6 & 28.5 & 12.3 & 18.8 & 30.9 & 40.1 & 24.3 & 39.0 & 41.8 & 37.3 & 32.7 \\
    \rowcolor{cyan!7} \textsc{Qwen2VL} \cite{wang2024qwen2} 2B & 56.4 & 27.0 & 13.5 & \textbf{32.8} & 43.7 & 50.6 & 27.8 & \textbf{57.4} & 47.9 & 42.7 & 40.0 \\
    \rowcolor{cyan!7} \textsc{Qwen2VL} \cite{wang2024qwen2} 7B & 55.8 & 28.5 & 20.7 & 20.6 & 41.8 & 50.6 & 25.1 & 48.5 & 48.1 & 43.2 & 38.3 \\
    \midrule
    \rowcolor{pink!25} \multicolumn{12}{l}{\footnotesize \textit{Open-world baselines}} \\
    \rowcolor{pink!25} CaSED~\cite{conti2023vocabulary} & 65.3 & \textbf{39.9} & 32.2 & 30.0 & \textbf{55.6} & \textbf{64.1} & \textbf{62.4} & 47.1 & \textbf{52.4} & \textbf{53.4} & \textbf{50.2} \\
    \rowcolor{pink!25} CLIP retrieval & 41.3 & 23.6 & 22.4 & 30.7 & 40.3 & 46.7 & 41.7 & 48.8 & 39.1 & 38.5 & 37.3 \\
    \midrule
    \rowcolor{gray!10}\multicolumn{12}{l}{\footnotesize \textcolor{gray}{\textit{Closed-world baselines}}} \\
    \rowcolor{gray!10}\textcolor{gray}{CLIP~\cite{radford2021clip}} & \textcolor{gray}{90.8} & \textcolor{gray}{69.9} & \textcolor{gray}{67.7} & \textcolor{gray}{66.7} & \textcolor{gray}{83.4} & \textcolor{gray}{93.7} & \textcolor{gray}{91.8} & \textcolor{gray}{80.5} & \textcolor{gray}{92.2} & \textcolor{gray}{83.3} & \textcolor{gray}{82.0} \\
    \rowcolor{gray!10}\textcolor{gray}{SigLIP~\cite{zhai2023siglip}} & \textcolor{gray}{97.8} & \textcolor{gray}{75.6} & \textcolor{gray}{63.1} & \textcolor{gray}{80.0} & \textcolor{gray}{92.0} & \textcolor{gray}{96.4} & \textcolor{gray}{96.8} & \textcolor{gray}{98.1} & \textcolor{gray}{83.1} & \textcolor{gray}{89.6} & \textcolor{gray}{87.3} \\
    \bottomrule
  \end{tabular}}
  \caption{Semantic similarity on ten datasets. Higher is better, \textbf{bold} indicates best.
  }
  \label{tab:semantic_similarity}
\end{table*}

%% file: tab/concept_similarity.tex
\begin{table*}
  \centering
  \resizebox{\linewidth}{!}{
  \begin{tabular}{l|cccccccccc|c}
    \toprule
    & \multicolumn{11}{c}{\textbf{Concept similarity}} \\
    \textbf{Model} & C101 & DTD & ESAT & FGVC & FLWR & FOOD & PETS & CARS & S397 & U101 & Avg. \\
    \midrule
    \rowcolor{cyan!7} \textsc{Idefics2} \cite{laurenccon2025idefics} 8B & 76.3 & 38.5 & 30.9 & 29.7 & 41.5 & 48.4 & 35.3 & 37.5 & 49.9 & 54.6 & 44.3 \\
    \textsc{InstructBLIP} \cite{dai2023instructblip} Vicuna 7B & 75.3 & 39.1 & 31.6 & 28.6 & 43.6 & 60.0 & 37.9 & 40.0 & 52.6 & 55.3 & 46.4 \\
    \rowcolor{cyan!7} \textsc{InternVL2} \cite{chen2024internvl, chen2024far} 2B & 75.7 & 48.0 & \textbf{52.9} & 36.8 & 49.5 & 60.8 & 41.9 & 50.9 & 65.1 & 59.4 & 54.1 \\
    \rowcolor{cyan!7} \textsc{InternVL2} \cite{chen2024internvl,chen2024far} 4B & 76.1 & 48.6 & 51.5 & 37.9 & 51.0 & 63.0 & 41.9 & 50.5 & 65.4 & 59.1 & 54.5 \\
    \rowcolor{cyan!7} \textsc{InternVL2} \cite{chen2024internvl,chen2024far} 8B & 78.7 & 49.7 & 49.1 & 42.5 & 56.9 & 67.1 & 46.0 & 56.2 & 69.2 & \textbf{62.9} & 57.8 \\
    \textsc{LLaVA-1.5} \cite{liu2023llava} 7B & 72.1 & 41.3 & 51.6 & 29.0 & 41.6 & 56.8 & 35.9 & 46.2 & 59.4 & 55.5 & 48.9 \\
    \textsc{LLaVA-NeXT} \cite{li2024llavaNext} (Mistral 7B) & 79.8 & \textbf{51.0} & 49.5 & 37.5 & 55.1 & 70.0 & 55.3 & 56.3 & 68.7 & 62.7 & 58.6 \\
    \textsc{LLaVA-NeXT} \cite{li2024llavaNext} (Vicuna 7B) & 79.0 & 50.1 & 50.8 & 37.1 & 51.3 & 65.8 & 42.4 & 55.0 & 67.4 & 61.8 & 56.1 \\
    \rowcolor{cyan!7} \textsc{LLaVa-OV} \cite{li2024llava-ov} (Qwen2 0.5B) & 77.8 & 45.1 & 39.9 & 30.6 & 42.4 & 50.0 & 37.5 & 43.5 & 56.7 & 55.9 & 47.9 \\
    \rowcolor{cyan!7} \textsc{LLaVa-OV} \cite{li2024llava-ov} (Qwen2 7B) & 79.1 & 47.0 & 41.0 & 29.4 & 51.7 & 41.9 & 37.8 & 35.4 & 44.9 & 43.3 & 45.1 \\
    \textsc{Phi-3-Vision} \cite{abdin2024phi} & 74.1 & 44.0 & 25.3 & 29.1 & 43.0 & 58.3 & 40.3 & 42.9 & 56.1 & 49.1 & 46.2 \\
    \rowcolor{cyan!7} \textsc{Qwen2VL} \cite{wang2024qwen2} 2B & 79.4 & 47.3 & 24.2 & \textbf{56.0} & 67.9 & 75.7 & 46.7 & \textbf{68.6} & 70.0 & 56.6 & \textbf{59.2} \\
    \rowcolor{cyan!7} \textsc{Qwen2VL} \cite{wang2024qwen2} 7B & \textbf{81.3} & 50.4 & 39.8 & 30.8 & \textbf{68.8} & \textbf{76.9} & 43.1 & 56.0 & \textbf{70.6} & 59.1 & 57.7 \\
    \midrule
    \rowcolor{pink!25} \multicolumn{12}{l}{\footnotesize \textit{Open-world baselines}} \\
    \rowcolor{pink!25} CaSED~\cite{conti2023vocabulary} & 65.9 & 39.8 & 32.2 & 29.9 & 55.6 & 66.5 & 62.9 & 47.1 & 53.7 & 55.1 & 50.9 \\
    \rowcolor{pink!25} CLIP retrieval & 63.9 & 38.1 & 37.8 & 50.7 & 62.3 & 67.8 & \textbf{66.1} & 61.5 & 57.3 & 54.4 & 56.0 \\
    \midrule
    \rowcolor{gray!10}\multicolumn{12}{l}{\footnotesize \textcolor{gray}{\textit{Closed-world baselines}}} \\
    \rowcolor{gray!10}\textcolor{gray}{CLIP~\cite{radford2021clip}} & \textcolor{gray}{90.8} & \textcolor{gray}{69.9} & \textcolor{gray}{67.7} & \textcolor{gray}{66.7} & \textcolor{gray}{83.4} & \textcolor{gray}{93.7} & \textcolor{gray}{91.8} & \textcolor{gray}{80.5} & \textcolor{gray}{92.2} & \textcolor{gray}{83.3} & \textcolor{gray}{82.0} \\
    \rowcolor{gray!10}\textcolor{gray}{SigLIP~\cite{zhai2023siglip}} & \textcolor{gray}{97.8} & \textcolor{gray}{75.6} & \textcolor{gray}{63.1} & \textcolor{gray}{80.0} & \textcolor{gray}{92.0} & \textcolor{gray}{96.4} & \textcolor{gray}{96.8} & \textcolor{gray}{98.1} & \textcolor{gray}{83.1} & \textcolor{gray}{89.6} & \textcolor{gray}{87.3} \\
    \bottomrule
  \end{tabular}}
  \caption{Concept similarity on ten datasets. Higher is better, \textbf{bold} indicates best.
  }
  \label{tab:concept_similarity}
\end{table*}

%% file: tab/average_elo_ranking.tex
\begin{table}
  \centering
  \resizebox{0.75\linewidth}{!}{
  \begin{tabular}{ccc}
    \toprule
    \multicolumn{3}{c}{\textbf{Average Elo ratings}} \\
    \textbf{Rank} & \textbf{Model} & \textbf{Rating} \\
    \midrule
    1 & \textsc{Qwen2VL} \cite{wang2024qwen2} 2B & 1037 \\
    2 & \textsc{Qwen2VL} \cite{wang2024qwen2} 7B & 1037 \\
    3 & \textsc{Phi-3-Vision} \cite{abdin2024phi} & 1029 \\
    4 & \textsc{LLaVA-NeXT} \cite{li2024llavaNext} (Mistral 7B) & 1018 \\
    5 & \textsc{LLaVA-NeXT} \cite{li2024llavaNext} (Vicuna 7B) & 1015 \\
    6 & \textsc{LLaVa-OV} \cite{li2024llava-ov} (Qwen2 7B) & 1014 \\
    7 & \textsc{LLaVa-OV} \cite{li2024llava-ov} (Qwen2 0.5B) & 1007 \\
    8 & \textsc{InternVL2} \cite{chen2024internvl,chen2024far} 8B & 1004 \\
    9 & \textsc{InternVL2} \cite{chen2024internvl,chen2024far} 4B & 994 \\
    10 & \textsc{InternVL2} \cite{chen2024internvl, chen2024far} 2B & 991 \\
    11 & \textsc{LLaVA-1.5} \cite{liu2023llava} 7B & 984 \\
    12 & \textsc{InstructBLIP} \cite{dai2023instructblip} Vicuna 7B & 943 \\
    13 & \textsc{Idefics2} \cite{laurenccon2025idefics} 8B & 924 \\
    \bottomrule
  \end{tabular}}
  \caption{Elo ratings on the ten datasets. Higher scores indicate comparatively better responses from the models.}
  \label{tab:elo_ranking_average}
\end{table}

%% file: tab/grouped_all_extended.tex
\begin{table*}
\centering
\resizebox{\linewidth}{!}{%
\begin{tabular}{lcccccccccccccccc}
\toprule
 & \multicolumn{4}{c}{\textbf{Prototypical}} & \multicolumn{4}{c}{\textbf{Non-prototypical}} & \multicolumn{4}{c}{\textbf{Fine-grained}} &  \multicolumn{4}{c}{\textbf{Very fine-grained}} \\
\cmidrule(lr){2-5} \cmidrule(lr){6-9} \cmidrule(lr){10-13} \cmidrule(lr){14-17}
\textbf{Model} & \textbf{TI} & \textbf{LI} & \textbf{SS} & \textbf{CS} & \textbf{TI} & \textbf{LI} & \textbf{SS} & \textbf{CS} & \textbf{TI} & \textbf{LI} & \textbf{SS} & \textbf{CS} & \textbf{TI} & \textbf{LI} & \textbf{SS} & \textbf{CS} \\
\midrule
\rowcolor{cyan!7} \textsc{Idefics2} \cite{laurenccon2025idefics} 8B & 30.8 & 52.7 & 54.5 & 63.1 & 3.7 & 27.9 & 35.4 & 41.3 & 3.0 & 49.9 & 38.0 & 41.7 & 0.0 & 67.0 & 29.6 & 33.6 \\
\textsc{InstructBLIP} \cite{dai2023instructblip} Vicuna 7B & 29.7 & 56.3 & 56.8 & 64.0 & 6.0 & 27.1 & 37.0 & 42.0 & 10.4 & 48.8 & 35.6 & 47.2 & 0.0 & 61.0 & 30.0 & 34.3 \\
\rowcolor{green!7}(*) \textsc{InstructBLIP} \cite{dai2023instructblip} Vicuna 13B & 22.7 & 47.7 & 49.5 & 57.8 & 4.4 & 27.3 & 34.2 & 41.2 & 6.6 & 36.7 & 30.2 & 41.4 & 0.0 & 52.9 & 31.5 & 34.2 \\
\rowcolor{cyan!7} \textsc{InternVL2} \cite{chen2024internvl, chen2024far} 2B & 36.9 & 69.9 & 46.9 & 70.4 & 10.2 & 45.2 & 31.6 & 53.4 & 14.9 & 47.0 & 31.6 & 50.7 & 0.7 & 32.9 & 33.1 & 43.9 \\
\rowcolor{cyan!7} \textsc{InternVL2} \cite{chen2024internvl, chen2024far} 4B & 36.3 & 68.5 & 46.5 & 70.8 & 10.1 & 42.1 & 30.8 & 53.1 & 16.2 & 44.4 & 32.0 & 52.0 & 1.7 & 36.8 & 33.8 & 44.2 \\
\rowcolor{cyan!7} \textsc{InternVL2} \cite{chen2024internvl, chen2024far} 8B & 40.6 & 74.4 & 48.2 & 74.0 & 11.0 & 46.2 & 31.9 & {53.9} & 22.3 & 46.7 & 34.8 & 56.7 & 2.3 & 32.5 & 36.0 & 49.4 \\
\rowcolor{green!7}(*) \textsc{InternVL2} \cite{chen2024internvl, chen2024far} 26B & 46.6 & 78.6 & 49.1 & \textbf{77.7} & \textbf{15.8} & \textbf{58.7} & 36.7 & \textbf{60.5} & 36.5 & 58.9 & 40.2 & 65.0 & 7.1 & 40.8 & 40.9 & 59.3 \\
\textsc{LLaVA-1.5} \cite{liu2023llava} 7B & 34.6 & 63.1 & 45.3 & 65.8 & 8.6 & 44.3 & 33.0 & 49.5 & 8.4 & 46.5 & 28.2 & 44.8 & 0.0 & 41.0 & 28.6 & 37.6 \\
\rowcolor{green!7}(*) \textsc{LLaVA-1.5} \cite{liu2023llava} 13B & 35.7 & 63.5 & 47.0 & 66.7 & 9.5 & 43.0 & 34.1 & 51.2 & 8.8 & 48.0 & 28.7 & 44.9 & 0.0 & 37.4 & 28.9 & 37.8 \\
\textsc{LLaVA-NeXT} \cite{li2024llavaNext} (Mistral 7B) & 41.7 & 73.9 & 45.9 & 74.3 & 11.3 & 46.8 & 31.2 & 54.4 & 26.8 & 43.7 & 35.3 & 60.1 & 1.4 & 47.2 & 34.2 & 46.9 \\
\textsc{LLaVA-NeXT} \cite{li2024llavaNext} (Vicuna 7B) & 39.5 & 72.8 & 46.2 & 73.2 & 10.6 & 45.9 & 31.1 & 54.2 & 16.9 & 44.5 & 32.2 & 53.2 & 1.3 & 42.2 & 34.5 & 46.1 \\
\rowcolor{green!7}(*) \textsc{LLaVA-NeXT} \cite{li2024llavaNext} (Vicuna 13B) & 42.2 & 73.6 & 46.2 & 75.3 & 11.4 & 46.5 & 32.4 & 55.5 & 26.2 & 44.0 & 36.1 & 60.4 & 1.3 & 33.4 & 34.4 & 47.0 \\
\rowcolor{green!7}(*) \textsc{LLaVA-NeXT} \cite{li2024llavaNext} (Yi 34B) & 39.2 & 74.9 & 46.2 & 73.9 & 12.2 & 49.3 & 33.1 & 56.6 & 25.3	& 43.1 & 35.1 & 60.0 & 0.9 & 42.5 & 33.5 & 45.3 \\
\rowcolor{cyan!7} \textsc{LLaVA-OV} \cite{li2024llava-ov} (Qwen2 0.5B) & 34.4 & 64.4 & 54.0 & 67.3 & 7.3 & 37.0 & 32.8 & 47.0 & 6.0 & 42.7 & 38.5 & 43.3 & 0.6 & 65.6 & 30.5 & 37.1 \\
\rowcolor{cyan!7} \textsc{LLaVA-OV} \cite{li2024llava-ov} (Qwen2 7B)  & 30.8 & 53.2 & 56.1 & 62.0 & 7.2 & 28.1 & 31.6 & 43.8 & 6.4 & 40.4 & 39.0 & 43.8 & 0.0 & 76.7 & 31.9 & 32.4 \\
\textsc{Phi-3-Vision} \cite{abdin2024phi} & 34.1 & 60.1 & 47.7 & 65.1 & 6.0 & 28.7 & 26.0 & 39.5 & 13.4 & 49.1 & 31.8 & 47.2 & 0.2 & 45.0 & 28.9 & 36.0 \\
\rowcolor{cyan!7} \textsc{Qwen2VL} \cite{wang2024qwen2} 2B & 44.9 & 77.8 & 52.2 & 74.7 & 7.8 & 34.3 & 27.7 & 42.7 & 35.7 & 62.5 & 40.7 & 63.4 & \textbf{12.9} & 60.7 & \textbf{45.1} & \textbf{62.3} \\
\rowcolor{cyan!7} \textsc{Qwen2VL} \cite{wang2024qwen2} 7B & 46.4 & \textbf{78.7} & 51.9 & 76.0 & 10.3 & 42.6 & 30.8 & 49.8 & {34.6} & 64.0 & 39.2 & 62.9 & 0.8 & \textbf{63.0} & 34.5 & 43.4 \\
\rowcolor{green!7}(*) \textsc{Qwen2VL} \cite{wang2024qwen2} 72B & \textbf{47.7} & 78.2 & 49.1 & 76.7 & 10.4 & 42.1 & 28.6 & 48.6 & \textbf{48.1} & \textbf{66.6} & 43.4 & \textbf{71.8} & 11.9 & 59.1 & 40.6 & 58.8 \\
\midrule
\rowcolor{pink!25} \multicolumn{17}{l}{\footnotesize \textit{Open-world baselines}} \\
\rowcolor{pink!25} \textsc{CaSED} \cite{conti2023vocabulary} & 24.5 & 46.3 & \textbf{58.9} & 59.8 & 5.4 & 18.6 & \textbf{41.8} & 42.4 & 27.4 & 46.6 & \textbf{60.7} & 61.7 & 0.7 & 47.1 & 38.5 & 38.5 \\
\rowcolor{pink!25} CLIP retrieval & 28.6 & 42.9 & 40.2 & 60.6 & 7.5 & 24.6 & 28.1 & 43.4 & 32.4 & 45.4 & 42.9 & 65.4 & 7.0 & 18.1 & 39.7 & 56.1 \\
\midrule
\rowcolor{gray!10} \multicolumn{17}{l}{\footnotesize \textcolor{gray}{\textit{Closed-world baselines}}} \\
\rowcolor{gray!10}\textcolor{gray}{CLIP \cite{radford2021clip}}  & \multicolumn{2}{c}{\textcolor{gray}{76.4}} & \multicolumn{2}{c}{\textcolor{gray}{91.5}} & \multicolumn{2}{c}{\textcolor{gray}{56.0}} & \multicolumn{2}{c}{\textcolor{gray}{73.6}} & \multicolumn{2}{c}{\textcolor{gray}{85.0}} & \multicolumn{2}{c}{\textcolor{gray}{89.6}} & \multicolumn{2}{c}{\textcolor{gray}{51.7}} & \multicolumn{2}{c}{\textcolor{gray}{73.6}} \\
\rowcolor{gray!10}\textcolor{gray}{SigLIP \cite{zhai2023siglip}} & \multicolumn{2}{c}{\textcolor{gray}{81.8}} & \multicolumn{2}{c}{\textcolor{gray}{90.5}} & \multicolumn{2}{c}{\textcolor{gray}{61.7}} & \multicolumn{2}{c}{\textcolor{gray}{76.1}} & \multicolumn{2}{c}{\textcolor{gray}{92.6}} & \multicolumn{2}{c}{\textcolor{gray}{95.1}} & \multicolumn{2}{c}{\textcolor{gray}{69.2}} & \multicolumn{2}{c}{\textcolor{gray}{89.1}} \\
\bottomrule
\end{tabular}
}
\vspace{-4pt}
\caption{OW results with larger models (in \inlineColorbox{green!7}{green}) averaged on the grouped datasets. TI stands for text inclusion, LI for Llama inclusion, SS for semantic similarity, and CS for concept similarity. Higher is better, \textbf{bold} indicates best.}
\vspace{-12pt}
\label{tab:grouped_all_extended}
\end{table*}

%% file: tab/larger_and_commercial_models.tex
\begin{table*}[ht]
\centering
\begin{minipage}{0.52\linewidth}
\centering
\resizebox{\linewidth}{!}{%
\begin{tabular}{lcccc}
\toprule
\textbf{Model} & \textbf{TI} & \textbf{LI} & \textbf{SS} & \textbf{CS} \\
\midrule
\rowcolor{cyan!7}\textsc{Idefics2} 8B & 12.5 & 45.7 & 42.6 & 49.2 \\
\textsc{InstructBLIP} Vicuna 7B & 13.4 & 38.3 & 37.4 & 50.2 \\
\rowcolor{green!7}(*) \textsc{InstructBLIP} Vicuna 13B & 10.0 & 38.3 & 37.4 & 45.2 \\
\rowcolor{cyan!7}\textsc{InternVL2} 2B & 19.5 & 54.4 & 34.9 & 54.9 \\
\rowcolor{cyan!7}\textsc{InternVL2} 4B & 18.9 & 51.5 & 34.4 & 55.3 \\
\rowcolor{cyan!7}\textsc{InternVL2} 8B & 22.9 & 54.7 & 36.3 & 58.8 \\
\rowcolor{green!7}(*) \textsc{InternVL2} 26B & 31.3 & 63.1 & 39.5 & 63.9 \\
\textsc{LLaVA-1.5} 7B & 14.7 & 50.8 & 32.2 & 49.3 \\
\rowcolor{green!7}(*) \textsc{LLaVA-1.5} 13B  & 15.7 & 52.9 & 33.1 & 50.1 \\
\textsc{LLaVA-NeXT} (Mistral 7B) & 25.9 & 51.6 & 35.7 & 60.8 \\
\textsc{LLaVA-NeXT} (Vicuna 7B) & 20.2 & 52.3 & 34.6 & 56.9 \\
\rowcolor{green!7}(*) \textsc{LLaVA-NeXT} (Vicuna 13B)  & 25.7 & 52.6 & 36.4 & 61.4 \\
\rowcolor{green!7}(*) \textsc{LLaVA-NeXT} (Yi 34B)  & 24.5 & 52.4 & 36.1 & 60.7 \\
\rowcolor{cyan!7}\textsc{LLaVA-OV} (Qwen2 0.5B) & 15.3 & 51.8 & 42.8 & 51.7 \\
\rowcolor{cyan!7}\textsc{LLaVA-OV} (Qwen2 7B) & 17.3 & 50.6 & \textbf{43.9} & 51.8 \\
\textsc{Phi-3-Vision} & 17.9 & 51.6 & 34.9 & 50.1 \\
\rowcolor{cyan!7}\textsc{Qwen2VL} 2B & 28.5 & 59.8 & 39.5 & 59.6 \\
\rowcolor{cyan!7}\textsc{Qwen2VL} 7B & 29.2 & 62.2 & 38.9 & 60.5 \\
\rowcolor{green!7}(*) \textsc{Qwen2VL} 72B & 36.7 & 64.0 & 40.2 & 66.3 \\
\bottomrule
\end{tabular}
}
\end{minipage}
\hfill
\begin{minipage}{0.46\linewidth}
\centering
\resizebox{\linewidth}{!}{%
\begin{tabular}{lcccc}
\toprule
\textbf{Model} & \textbf{TI} & \textbf{LI} & \textbf{SS} & \textbf{CS} \\
\midrule
\rowcolor{red!25} \multicolumn{5}{l}{\footnotesize \textit{Reasoning models}} \\
\rowcolor{red!25} \textsc{InternVL2.5} 2B & 20.3 & 51.7 & 33.1 & 54.6 \\
\rowcolor{red!25} \textsc{InternVL2.5} 4B & 21.6 & 54.4 & 35.7 & 55.8 \\
\rowcolor{red!25} \textsc{InternVL2.5} 8B & 21.3 & 55.9 & 36.0 & 56.3 \\
\rowcolor{red!25} \textsc{Qwen2.5VL} 3B & 36.5 & 64.2 & 39.8 & 66.1 \\
\rowcolor{red!25} \textsc{Qwen2.5VL} 7B & \textbf{45.0} & 71.5 & 41.9 & \textbf{72.6} \\
\midrule
\rowcolor{ultraviolet!25} \multicolumn{5}{l}{\footnotesize \textit{Commercial models}} \\
\rowcolor{ultraviolet!25} (*) \textsc{GPT-4o-mini} & 29.5 & 70.3 & 39.9 & 63.1 \\
\rowcolor{ultraviolet!25} (*) \textsc{GPT-4o} & 27.4 & 66.3 & 41.2 & 59.9 \\
\rowcolor{ultraviolet!25} (*) \textsc{Claude Haiku 3.5} & 37.2 & 74.7 & 42.1 & 70.1 \\
\rowcolor{ultraviolet!25} (*) \textsc{Claude Sonnet 3.5} & 39.0 & \textbf{77.3} & 42.4 & 72.2 \\
\rowcolor{ultraviolet!25} (*) \textsc{Gemini 2.0 Flash} & 29.2 & 62.2 & 39.1 & 60.1 \\
\midrule
\rowcolor{pink!25} \multicolumn{5}{l}{\footnotesize \textit{Open-world baselines}} \\
\rowcolor{pink!25} \textsc{CaSED} & 22.3 & 42.2 & 55.3 & 55.9 \\
\rowcolor{pink!25} CLIP retrieval & 25.9 & 43.4 & 37.0 & 57.0 \\
\midrule
\rowcolor{gray!10} \multicolumn{5}{l}{\footnotesize \textcolor{gray}{\textit{Closed-world baselines}}} \\
\rowcolor{gray!10}\textcolor{gray}{CLIP}  & \multicolumn{2}{c}{\textcolor{gray}{75.5}} & \multicolumn{2}{c}{\textcolor{gray}{83.8}} \\
\rowcolor{gray!10}\textcolor{gray}{SigLIP} & \multicolumn{2}{c}{\textcolor{gray}{84.0}} & \multicolumn{2}{c}{\textcolor{gray}{90.4}} \\
\bottomrule
\end{tabular}
}
\end{minipage}
\caption{Results with larger (in \inlineColorbox{green!7}{green}) and commercial (in \inlineColorbox{ultraviolet!25}{purple}) models averaged on 5 datasets, \ie, Caltech101, DTD, Flowers102, OxfordPets, UCF101. TI stands for text inclusion, LI for Llama inclusion, SS for semantic similarity, and
CS for concept similarity. Higher is better, \textbf{bold} is best.
}
\label{tab:larger_and_commercial}
\end{table*}

%% file: tab/list_caption_describe.tex
\begin{table*}
  \centering
  \resizebox{\linewidth}{!}{
\begin{tabular}{l|ccc|ccc|ccc|ccc|ccc}
    \toprule
     & \multicolumn{3}{c}{\textbf{Caltech101}} & \multicolumn{3}{c}{\textbf{DTD}} & \multicolumn{3}{c}{\textbf{Flowers102}}  & \multicolumn{3}{c}{\textbf{OxfordPets}}  & \multicolumn{3}{c}{\textbf{UCF101}}  \\
     \textbf{Model} & \textbf{TI} & \textbf{LI} & \textbf{CS} & \textbf{TI} & \textbf{LI} & \textbf{CS} & \textbf{TI} & \textbf{LI} & \textbf{CS} & \textbf{TI} & \textbf{LI} & \textbf{CS} & \textbf{TI} & \textbf{LI} & \textbf{CS} \\
    \midrule
    \multicolumn{16}{l}{\textbf{List}} \\
    \textsc{Idefics2} \cite{laurenccon2025idefics} 8B & \ccc{-1.5} & \ccc{-12.8} & \ccc{-2.8} & \ccc{+3.6} & \ccc{+3.4} & \ccc{+2.0} & \ccc{+3.6} & \ccc{-19.3} & \ccc{-0.3} & \ccc{+1.8} & \ccc{-17.6} & \ccc{+3.5} & \ccc{+2.7} & \ccc{-6.6} & \ccc{-1.2} \\
    \textsc{InstructBLIP} \cite{dai2023instructblip} Vicuna 7B & \ccc{+12.5} & \ccc{-3.5} & \ccc{+4.5} & \ccc{+6.8} & \ccc{+15.8} & \ccc{+6.0} & \ccc{+13.7} & \ccc{-24.2} & \ccc{+5.9} & \ccc{+2.6} & \ccc{-29.1} & \ccc{+0.9} & \ccc{+6.7} & \ccc{+21.1} & \ccc{+9.3} \\
    \textsc{InternVL2} \cite{chen2024internvl, chen2024far} 2B & \ccc{-0.8} & \ccc{-6.1} & \ccc{+1.2} & \ccc{-1.5} & \ccc{-2.8} & \ccc{-1.7} & \ccc{-4.2} & \ccc{-23.5} & \ccc{-2.4} & \ccc{+4.8} & \ccc{-32.8} & \ccc{+2.2} & \ccc{-3.1} & \ccc{-11.2} & \ccc{-1.6} \\
    \textsc{InternVL2} \cite{chen2024internvl,chen2024far} 4B &\ccc{+1.8} & \ccc{-4.3} & \ccc{+0.7} & \ccc{-1.8} & \ccc{+3.6} & \ccc{-1.6} & \ccc{-3.7} & \ccc{-12.7} & \ccc{-5.7} & \ccc{+3.4} & \ccc{-35.5} & \ccc{+3.1} & \ccc{+2.2} & \ccc{-0.2} & \ccc{+0.6} \\
    \textsc{InternVL2} \cite{chen2024internvl,chen2024far} 8B &\ccc{-1.7} & \ccc{-2.2} & \ccc{-0.2} & \ccc{-0.8} & \ccc{+0.1} & \ccc{-1.2} & \ccc{-4.8} & \ccc{-2.3} & \ccc{-5.2} & \ccc{+2.3} & \ccc{-26.5} & \ccc{+2.4} & \ccc{+1.0} & \ccc{-3.9} & \ccc{-3.2} \\
\end{tabular}}
  \caption{Relative performance variation with multi-label prompts on five datasets. TI stands for text inclusion, LI for Llama inclusion, and CS for concept similarity.}
  \label{tab:list_caption_describe}
\end{table*}

%% file: tab/generic_specific.tex
\begin{table*}
  \centering
  \resizebox{\linewidth}{!}{
  \begin{tabular}{l|ccc|ccc|ccc|ccc|ccc|ccc}
    \toprule
     & \multicolumn{3}{c}{\textbf{DTD}} &  \multicolumn{3}{c}{\textbf{FGVCAircraft}} & \multicolumn{3}{c}{\textbf{Flowers102}} &  \multicolumn{3}{c}{\textbf{Food101}} & \multicolumn{3}{c}{\textbf{OxfordPets}} & \multicolumn{3}{c}{\textbf{StanfordCars}}  \\
     \textbf{Model} & \textbf{TI} & \textbf{LI} & \textbf{CS} & \textbf{TI} & \textbf{LI} & \textbf{CS} & \textbf{TI} & \textbf{LI} & \textbf{CS} & \textbf{TI} & \textbf{LI} & \textbf{CS} & \textbf{TI} & \textbf{LI} & \textbf{CS} & \textbf{TI} & \textbf{LI} & \textbf{CS} \\
    \midrule
    \multicolumn{16}{l}{\textbf{Be generic}} \\
    \textsc{Idefics2} \cite{laurenccon2025idefics} 8B & \ccc{-7.0} & \ccc{-9.7} & \ccc{-8.5} & \ccc{-2.4} & \ccc{+27.2} & \ccc{-2.4} & \ccc{-25.7} & \ccc{+8.9} & \ccc{-13.3} & \ccc{-19.8} & \ccc{+30.5} & \ccc{-20.7} & \ccc{-0.5} & \ccc{-15.5} & \ccc{-4.2} & \ccc{+0.0} & \ccc{+8.2} & \ccc{-19.1} \\
    \textsc{InstructBLIP} \cite{dai2023instructblip} Vicuna 7B & \ccc{-10.6} & \ccc{-20.7} & \ccc{-9.3} & \ccc{-0.7} & \ccc{-9.1} & \ccc{-4.4} & \ccc{-29.6} & \ccc{+2.5} & \ccc{-19.0} & \ccc{-15.7} & \ccc{-13.9} & \ccc{-13.3} & \ccc{-13.6} & \ccc{-20.5} & \ccc{-8.4} & \ccc{+0.0} & \ccc{+10.2} & \ccc{-24.5} \\
    \textsc{InternVL2} \cite{chen2024internvl, chen2024far} 2B & \ccc{-3.0} & \ccc{+3.8} & \ccc{-3.9} & \ccc{-2.5} & \ccc{+43.5} & \ccc{-15.8} & \ccc{-6.1} & \ccc{+31.8} & \ccc{-11.4} & \ccc{-4.5} & \ccc{-11.5} & \ccc{-6.2} & \ccc{-14.8} & \ccc{-8.2} & \ccc{-9.1} & \ccc{+0.0} & \ccc{+13.9} & \ccc{-8.2} \\
    \textsc{InternVL2} \cite{chen2024internvl,chen2024far} 4B & \ccc{-6.3} & \ccc{-4.7} & \ccc{-3.9} & \ccc{-7.2} & \ccc{+63.2} & \ccc{-20.1} & \ccc{-13.7} & \ccc{+35.8} & \ccc{-15.2} & \ccc{-13.3} & \ccc{-11.6} & \ccc{-10.4} & \ccc{-10.0} & \ccc{-3.6} & \ccc{-7.1} & \ccc{-0.0} & \ccc{+51.4} & \ccc{-14.1} \\
\bottomrule
  \end{tabular}}
  \caption{Relative performance variation with the generic/specific prompts on six datasets. TI stands for text inclusion, LI for Llama inclusion, and CS for concept similarity.}
  \label{tab:more_generic_more_specific_relative}
\end{table*}

%% file: tab/fine_grained.tex
\begin{table*}
  \centering
  \resizebox{\linewidth}{!}{
  \begin{tabular}{l|ccc|ccc|ccc|ccc|ccc|ccc}
    \toprule
     & \multicolumn{3}{c}{\textbf{DTD}} &  \multicolumn{3}{c}{\textbf{FGVCAircraft}} & \multicolumn{3}{c}{\textbf{Flowers102}} &  \multicolumn{3}{c}{\textbf{Food101}} & \multicolumn{3}{c}{\textbf{OxfordPets}} & \multicolumn{3}{c}{\textbf{StanfordCars}}  \\
     \textbf{Model} & \textbf{TI} & \textbf{LI} & \textbf{CS} & \textbf{TI} & \textbf{LI} & \textbf{CS} & \textbf{TI} & \textbf{LI} & \textbf{CS} & \textbf{TI} & \textbf{LI} & \textbf{CS} & \textbf{TI} & \textbf{LI} & \textbf{CS} & \textbf{TI} & \textbf{LI} & \textbf{CS} \\
    \midrule
    \textsc{Idefics2} \cite{laurenccon2025idefics} 8B & \ccc{+6.0} & \ccc{+3.2} & \ccc{+6.8} & \ccc{+2.4} & \ccc{-37.3} & \ccc{+2.2} & \ccc{+24.9} & \ccc{-10.2} & \ccc{+12.9} & \ccc{+15.3} & \ccc{-19.6} & \ccc{+15.0} & \ccc{+0.4} & \ccc{+9.9} & \ccc{+3.2} & \ccc{+0.0} & \ccc{-2.1} & \ccc{+16.8} \\
    \textsc{InstructBLIP} \cite{dai2023instructblip} Vicuna 7B & \ccc{+9.0} & \ccc{+11.2} & \ccc{+6.1} & \ccc{+0.7} & \ccc{-0.3} & \ccc{+4.1} & \ccc{+26.6} & \ccc{-11.1} & \ccc{+17.9} & \ccc{+6.4} & \ccc{+1.5} & \ccc{+5.7} & \ccc{+12.8} & \ccc{+16.6} & \ccc{+7.9} & \ccc{+0.0} & \ccc{-7.4} & \ccc{+19.9} \\
    \textsc{InternVL2} \cite{chen2024internvl, chen2024far} 2B & \ccc{+2.3} & \ccc{-5.9} & \ccc{+0.6} & \ccc{+1.5} & \ccc{-20.9} & \ccc{+7.6} & \ccc{+2.2} & \ccc{-24.6} & \ccc{+7.1} & \ccc{+0.8} & \ccc{+5.4} & \ccc{+1.8} & \ccc{+9.8} & \ccc{+5.9} & \ccc{+6.5} & \ccc{-0.0} & \ccc{-8.4} & \ccc{+3.1} \\
    \textsc{InternVL2} \cite{chen2024internvl,chen2024far} 4B & \ccc{+3.8} & \ccc{+0.9} & \ccc{+1.3} & \ccc{+3.8} & \ccc{-25.3} & \ccc{+10.2} & \ccc{+5.8} & \ccc{-12.0} & \ccc{+5.6} & \ccc{+3.0} & \ccc{+4.1} & \ccc{+2.2} & \ccc{+4.7} & \ccc{+3.6} & \ccc{+3.8} & \ccc{+0.0} & \ccc{-18.5} & \ccc{+4.9} \\
\bottomrule
  \end{tabular}}
  \caption{Relative performance variation with dataset-specific prompts on six datasets. TI stands for text inclusion, LI for Llama inclusion, and CS for concept similarity.}
  \label{tab:fine_grained_relative}
\end{table*}

%% file: tab/group_errors_generic_specific.tex
\begin{table}
  \centering
  \resizebox{\linewidth}{!}{ 
  \begin{tabular}{l|cc|cc}
    \toprule
     & \multicolumn{2}{c|}{\textbf{Correct}} & \multicolumn{2}{c}{\textbf{Wrong}} \\
     \textbf{Model} & \textbf{Specific} & \textbf{Generic} & \textbf{Specific} & \textbf{Generic} \\
    \midrule
    \multicolumn{5}{l}{\textbf{Be generic}} \\
    \textsc{Idefics2}     \cite{laurenccon2025idefics} 8B & \ccc{-5.9} & \ccc{+11.3} & \ccc{-1.4} & \ccc{-4.0} \\
    \textsc{InstructBLIP} \cite{dai2023instructblip} Vicuna 7B & \ccc{-9.9} & \ccc{+2.2} & \ccc{+0.5} & \ccc{+7.3} \\
    \textsc{InternVL2} \cite{chen2024internvl, chen2024far} 2B & \ccc{-5.3} & \ccc{+4.8} & \ccc{-1.2} & \ccc{+1.7} \\
    \textsc{InternVL2} \cite{chen2024internvl,chen2024far} 4B & \ccc{-9.3} & \ccc{+14.5} & \ccc{-3.9} & \ccc{-1.4} \\
    \textsc{InternVL2} \cite{chen2024internvl,chen2024far} 8B & \ccc{-20.6} & \ccc{+21.9} & \ccc{-5.5} & \ccc{+4.2} \\
\bottomrule
  \end{tabular}
  }
  \caption{Gains on the types of model prediction when instructing the models to be more generic/specific, and when using dataset-specific prompts techniques on six datasets, \ie, DTD, FGVCAircraft, Flowers102, Food101, OxfordPets, StanfordCars.}
  \label{tab:group_errors_more_generic}
\end{table}

%% file: tab/test_time.tex
\begin{table*}
  \centering
  \resizebox{\linewidth}{!}{
  \begin{tabular}{l|ccc|ccc|ccc|ccc|ccc}
    \toprule
     & \multicolumn{3}{c}{\textbf{Caltech101}} & \multicolumn{3}{c}{\textbf{DTD}} & \multicolumn{3}{c}{\textbf{Flowers102}}  & \multicolumn{3}{c}{\textbf{OxfordPets}}  & \multicolumn{3}{c}{\textbf{UCF101}}  \\
     \textbf{Model} & \textbf{TI} & \textbf{LI} & \textbf{CS} & \textbf{TI} & \textbf{LI} & \textbf{CS} & \textbf{TI} & \textbf{LI} & \textbf{CS} & \textbf{TI} & \textbf{LI} & \textbf{CS} & \textbf{TI} & \textbf{LI} & \textbf{CS} \\
    \midrule
    \multicolumn{16}{l}{\textbf{Zero-shot chain-of-thought}} \\
    \textsc{InternVL2} \cite{chen2024internvl, chen2024far} 2B & \ccc{+0.3} & \ccc{-0.5} & \ccc{+4.0} & \ccc{+2.3} & \ccc{+18.5} & \ccc{+5.3} & \ccc{-3.5} & \ccc{-7.8} & \ccc{+1.3} & \ccc{+4.9} & \ccc{-14.9} & \ccc{+5.3} & \ccc{+6.3} & \ccc{+5.3} & \ccc{+6.3} \\
    \textsc{InternVL2} \cite{chen2024internvl, chen2024far} 4B & \ccc{-5.1} & \ccc{+2.2} & \ccc{-0.7} & \ccc{+1.4} & \ccc{+29.3} & \ccc{+3.1} & \ccc{+0.9} & \ccc{+16.0} & \ccc{+2.7} & \ccc{+3.0} & \ccc{+6.9} & \ccc{+4.9} & \ccc{+4.6} & \ccc{+23.9} & \ccc{+4.6} \\
    \textsc{InternVL2} \cite{chen2024internvl, chen2024far} 8B & \ccc{-2.3} & \ccc{+2.2} & \ccc{+0.5} & \ccc{+4.0} & \ccc{+20.7} & \ccc{+3.5} & \ccc{-1.7} & \ccc{+13.8} & \ccc{+0.1} & \ccc{+2.3} & \ccc{-9.7} & \ccc{+4.2} & \ccc{+4.3} & \ccc{+14.9} & \ccc{+3.2} \\
    \textsc{Qwen2VL} \cite{wang2024qwen2} 2B & \ccc{+1.8} & \ccc{+3.6} & \ccc{+5.5} & \ccc{+6.5} & \ccc{+25.4} & \ccc{+8.2} & \ccc{-1.9} & \ccc{+5.4} & \ccc{+2.0} & \ccc{+6.8} & \ccc{+3.6} & \ccc{+6.7} & \ccc{+5.7} & \ccc{+23.2} & \ccc{+7.9} \\
    \textsc{Qwen2VL} \cite{wang2024qwen2} 7B & \ccc{-2.9} & \ccc{+5.0} & \ccc{+1.5} & \ccc{+6.5} & \ccc{+21.5} & \ccc{+5.5} & \ccc{-4.9} & \ccc{+13.6} & \ccc{-1.5} & \ccc{+7.1} & \ccc{+17.0} & \ccc{+7.7} & \ccc{+2.3} & \ccc{+23.9} & \ccc{+5.4} \\
\end{tabular}}
  \caption{Relative performance variation with chain-of-thought prompts on five datasets. TI stands for text inclusion, LI for Llama inclusion, and CS for concept similarity.}
  \label{tab:test_time}
\end{table*}

%% file: tab/group_errors_chain_of_thought.tex
\begin{table}
  \centering
  \resizebox{\linewidth}{!}{ 
  \begin{tabular}{l|cc|cc}
    \toprule
     & \multicolumn{2}{c|}{\textbf{Correct}} & \multicolumn{2}{c}{\textbf{Wrong}} \\
     \textbf{Model} & \textbf{Specific} & \textbf{Generic} & \textbf{Specific} & \textbf{Generic} \\
    \midrule
    \multicolumn{5}{l}{\textbf{Zero-shot chain-of-thought}} \\
    \textsc{InternVL2} \cite{chen2024internvl, chen2024far} 2B & \ccc{+3.5} & \ccc{-5.5} & \ccc{+1.4} & \ccc{+0.6} \\
    \textsc{InternVL2} \cite{chen2024internvl,chen2024far} 4B & \ccc{+4.8} & \ccc{+9.7} & \ccc{-2.0} & \ccc{-12.5} \\
    \textsc{InternVL2} \cite{chen2024internvl,chen2024far} 8B & \ccc{+3.9} & \ccc{+2.1} & \ccc{-1.1} & \ccc{-4.9} \\
    \textsc{Qwen2VL} \cite{wang2024qwen2} 2B & \ccc{+8.0} & \ccc{+3.1} & \ccc{-0.4} & \ccc{-10.8} \\
    \textsc{Qwen2VL} \cite{wang2024qwen2} 7B & \ccc{+6.9} & \ccc{+8.6} & \ccc{-2.5} & \ccc{-13.1} \\
\bottomrule
  \end{tabular}
  }
  \caption{
  Gains on the types of model prediction when instructing the models to reason with chain-of-thought, and when using reasoning models on five datasets, \ie, Caltech101, DTD, Flowers102, OxfordPets, UCF101.
  }
  \label{tab:group_errors_test_time}
\end{table}

%% file: tab/group_errors_list_caption_describe.tex
\begin{table}
  \centering
  \resizebox{\linewidth}{!}{ 
  \begin{tabular}{l|cc|cc}
    \toprule
     & \multicolumn{2}{c|}{\textbf{Correct}} & \multicolumn{2}{c}{\textbf{Wrong}} \\
     \textbf{Model} & \textbf{Specific} & \textbf{Generic} & \textbf{Specific} & \textbf{Generic} \\
    \midrule
    \multicolumn{5}{l}{\textbf{List}} \\
    \textsc{Idefics2}     \cite{laurenccon2025idefics} 8B & \ccc{-4.2} & \ccc{-9.5} & \ccc{+4.0} & \ccc{+9.7} \\
    \textsc{InstructBLIP} \cite{dai2023instructblip} Vicuna 7B & \ccc{+9.6} & \ccc{-16.1} & \ccc{+1.5} & \ccc{+5.0} \\
    \textsc{InternVL2} \cite{chen2024internvl, chen2024far} 2B & \ccc{-3.1} & \ccc{-14.2} & \ccc{+3.5} & \ccc{+13.9} \\
    \textsc{InternVL2} \cite{chen2024internvl,chen2024far} 4B & \ccc{-1.4} & \ccc{-10.9} & \ccc{+0.9} & \ccc{+11.3} \\
    \textsc{InternVL2} \cite{chen2024internvl,chen2024far} 8B & \ccc{-2.6} & \ccc{-6.7} & \ccc{+0.8} & \ccc{+8.5} \\
\bottomrule
  \end{tabular}
  }
  \caption{Gains on the types of model prediction when instructing the models with multi-label prompts on five datasets, \ie, Caltech101, DTD, Flowers102, OxfordPets, UCF101.}
  \label{tab:group_errors_list_caption_describe}
\end{table}

%% file: tab/reasoning_models.tex
\begin{table*}
  \centering
  \resizebox{\linewidth}{!}{
  \begin{tabular}{l|cccccccccc|c}
    \toprule
    & \multicolumn{11}{c}{\textbf{Datasets}} \\
    \textbf{Model} & C101 & DTD & ESAT & FGVC & FLWR & FOOD & PETS & CARS & S397 & U101 & Avg. \\
    \midrule
    \multicolumn{12}{l}{\textbf{Text inclusion}} \\
    \textsc{InternVL2.5} \cite{chen2024expanding} 2B & 55.8 & 12.6 & \textbf{12.6} & 1.5 & 10.9 & 17.0 & 8.7 & 0.0 & 16.3 & 13.7 & 14.9 \\
    \textsc{InternVL2.5} \cite{chen2024expanding} 4B & 55.6 & 10.9 & 12.1 & 0.9 & 12.2 & 24.9 & 14.6 & 0.0 & 23.7 & 14.9 & 17.0 \\
    \textsc{InternVL2.5} \cite{chen2024expanding} 8B & 56.4 & 12.1 & 8.4 & 3.0 & 16.8 & 29.7 & 7.2 & \textbf{0.1} & 24.7 & 13.8 & 17.2 \\
    \textsc{Qwen2.5VL} \cite{bai2025qwen2.5} 3B & 62.1 & 13.9 & 1.6 & 18.8 & 49.7 & 44.2 & 38.9 & 0.0 & 30.7 & 18.0 & 27.8 \\
    \textsc{Qwen2.5VL} \cite{bai2025qwen2.5} 7B & \textbf{65.6} & \textbf{16.7} & 4.4 & \textbf{32.7} & \textbf{56.1} & \textbf{54.9} & \textbf{65.1} & 0.0 & \textbf{33.6} & \textbf{21.5} & \textbf{35.1} \\
\bottomrule
  \end{tabular}}
  \caption{OW results of reasoning models on ten datasets. Higher is better, \textbf{bold} indicates best. Note that the Llama inclusion for CLIP closed-world equals the textual inclusion scores.}
  \label{tab:reasoning_models}
\end{table*}

%% file: main.bbl
\begin{thebibliography}{74}
\providecommand{\natexlab}[1]{#1}
\providecommand{\url}[1]{\texttt{#1}}
\expandafter\ifx\csname urlstyle\endcsname\relax
  \providecommand{\doi}[1]{doi: #1}\else
  \providecommand{\doi}{doi: \begingroup \urlstyle{rm}\Url}\fi

\bibitem[Abdin et~al.(2024)Abdin, Aneja, Awadalla, Awadallah, Awan, Bach, Bahree, Bakhtiari, Bao, Behl, et~al.]{abdin2024phi}
Marah Abdin, Jyoti Aneja, Hany Awadalla, Ahmed Awadallah, Ammar~Ahmad Awan, Nguyen Bach, Amit Bahree, Arash Bakhtiari, Jianmin Bao, Harkirat Behl, et~al.
\newblock Phi-3 technical report: A highly capable language model locally on your phone.
\newblock \emph{arXiv preprint arXiv:2404.14219}, 2024.

\bibitem[Alayrac et~al.(2022)Alayrac, Donahue, Luc, Miech, Barr, Hasson, Lenc, Mensch, Millican, Reynolds, et~al.]{alayrac2022flamingo}
Jean-Baptiste Alayrac, Jeff Donahue, Pauline Luc, Antoine Miech, Iain Barr, Yana Hasson, Karel Lenc, Arthur Mensch, Katherine Millican, Malcolm Reynolds, et~al.
\newblock Flamingo: a visual language model for few-shot learning.
\newblock \emph{NeurIPS}, 2022.

\bibitem[Bai et~al.(2023)Bai, Bai, Yang, Wang, Tan, Wang, Lin, Zhou, and Zhou]{bai2023qwen}
Jinze Bai, Shuai Bai, Shusheng Yang, Shijie Wang, Sinan Tan, Peng Wang, Junyang Lin, Chang Zhou, and Jingren Zhou.
\newblock Qwen-vl: A frontier large vision-language model with versatile abilities.
\newblock \emph{arXiv preprint arXiv:2308.12966}, 2023.

\bibitem[Bai et~al.(2025)Bai, Chen, Liu, Wang, Ge, Song, Dang, Wang, Wang, Tang, et~al.]{bai2025qwen2.5}
Shuai Bai, Keqin Chen, Xuejing Liu, Jialin Wang, Wenbin Ge, Sibo Song, Kai Dang, Peng Wang, Shijie Wang, Jun Tang, et~al.
\newblock Qwen2.5-vl technical report.
\newblock \emph{arXiv preprint arXiv:2502.13923}, 2025.

\bibitem[Belouadah et~al.(2021)Belouadah, Popescu, and Kanellos]{belouadah2021ILsurvey}
Eden Belouadah, Adrian Popescu, and Ioannis Kanellos.
\newblock A comprehensive study of class incremental learning algorithms for visual tasks.
\newblock \emph{Neural Networks}, 135:\penalty0 38--54, 2021.

\bibitem[Bendale and Boult(2015)]{bendale2015towards-open-world}
Abhijit Bendale and Terrance Boult.
\newblock Towards open world recognition.
\newblock In \emph{CVPR}, 2015.

\bibitem[Beyer et~al.(2020)Beyer, H{\'e}naff, Kolesnikov, Zhai, and Oord]{beyer2020we}
Lucas Beyer, Olivier~J H{\'e}naff, Alexander Kolesnikov, Xiaohua Zhai, and A{\"a}ron van~den Oord.
\newblock Are we done with imagenet?
\newblock \emph{arXiv preprint arXiv:2006.07159}, 2020.

\bibitem[Bossard et~al.(2014)Bossard, Guillaumin, and Van~Gool]{bossard2014food}
Lukas Bossard, Matthieu Guillaumin, and Luc Van~Gool.
\newblock Food-101--mining discriminative components with random forests.
\newblock In \emph{ECCV}, 2014.

\bibitem[Chen et~al.(2024{\natexlab{a}})Chen, Chen, Zhang, Wang, Liu, Zhou, Zhang, Wan, Zhou, and Sun]{chen2024mllmasajudge}
Dongping Chen, Ruoxi Chen, Shilin Zhang, Yaochen Wang, Yinuo Liu, Huichi Zhou, Qihui Zhang, Yao Wan, Pan Zhou, and Lichao Sun.
\newblock Mllm-as-a-judge: Assessing multimodal llm-as-a-judge with vision-language benchmark.
\newblock In \emph{ICML}, 2024{\natexlab{a}}.

\bibitem[Chen et~al.(2024{\natexlab{b}})Chen, Wang, Cao, Liu, Gao, Cui, Zhu, Ye, Tian, Liu, et~al.]{chen2024expanding}
Zhe Chen, Weiyun Wang, Yue Cao, Yangzhou Liu, Zhangwei Gao, Erfei Cui, Jinguo Zhu, Shenglong Ye, Hao Tian, Zhaoyang Liu, et~al.
\newblock Expanding performance boundaries of open-source multimodal models with model, data, and test-time scaling.
\newblock \emph{arXiv preprint arXiv:2412.05271}, 2024{\natexlab{b}}.

\bibitem[Chen et~al.(2024{\natexlab{c}})Chen, Wang, Cao, Liu, Gao, Cui, Zhu, Ye, Tian, Liu, et~al.]{chen2024internvl2.5}
Zhe Chen, Weiyun Wang, Yue Cao, Yangzhou Liu, Zhangwei Gao, Erfei Cui, Jinguo Zhu, Shenglong Ye, Hao Tian, Zhaoyang Liu, et~al.
\newblock Expanding performance boundaries of open-source multimodal models with model, data, and test-time scaling.
\newblock \emph{arXiv preprint arXiv:2412.05271}, 2024{\natexlab{c}}.

\bibitem[Chen et~al.(2024{\natexlab{d}})Chen, Wang, Tian, Ye, Gao, Cui, Tong, Hu, Luo, Ma, et~al.]{chen2024far}
Zhe Chen, Weiyun Wang, Hao Tian, Shenglong Ye, Zhangwei Gao, Erfei Cui, Wenwen Tong, Kongzhi Hu, Jiapeng Luo, Zheng Ma, et~al.
\newblock How far are we to gpt-4v? closing the gap to commercial multimodal models with open-source suites.
\newblock \emph{Science China Information Sciences}, 67\penalty0 (12):\penalty0 220101, 2024{\natexlab{d}}.

\bibitem[Chen et~al.(2024{\natexlab{e}})Chen, Wu, Wang, Su, Chen, Xing, Zhong, Zhang, Zhu, Lu, et~al.]{chen2024internvl}
Zhe Chen, Jiannan Wu, Wenhai Wang, Weijie Su, Guo Chen, Sen Xing, Muyan Zhong, Qinglong Zhang, Xizhou Zhu, Lewei Lu, et~al.
\newblock Internvl: Scaling up vision foundation models and aligning for generic visual-linguistic tasks.
\newblock In \emph{CVPR}, 2024{\natexlab{e}}.

\bibitem[Chiang et~al.(2023)Chiang, Li, Lin, Sheng, Wu, Zhang, Zheng, Zhuang, Zhuang, Gonzalez, et~al.]{chiang2023vicuna}
Wei-Lin Chiang, Zhuohan Li, Ziqing Lin, Ying Sheng, Zhanghao Wu, Hao Zhang, Lianmin Zheng, Siyuan Zhuang, Yonghao Zhuang, Joseph~E Gonzalez, et~al.
\newblock Vicuna: An open-source chatbot impressing gpt-4 with 90\%* chatgpt quality.
\newblock \emph{See https://vicuna.lmsys.org (accessed 14 April 2023)}, 2023.

\bibitem[Chiang et~al.(2024)Chiang, Zheng, Sheng, Angelopoulos, Li, Li, Zhu, Zhang, Jordan, Gonzalez, et~al.]{chiang2024chatbot}
Wei-Lin Chiang, Lianmin Zheng, Ying Sheng, Anastasios~Nikolas Angelopoulos, Tianle Li, Dacheng Li, Banghua Zhu, Hao Zhang, Michael Jordan, Joseph~E Gonzalez, et~al.
\newblock Chatbot arena: An open platform for evaluating llms by human preference.
\newblock In \emph{ICML}, 2024.

\bibitem[Cimpoi et~al.(2014)Cimpoi, Maji, Kokkinos, Mohamed, and Vedaldi]{cimpoi2014dtd}
Mircea Cimpoi, Subhransu Maji, Iasonas Kokkinos, Sammy Mohamed, and Andrea Vedaldi.
\newblock Describing textures in the wild.
\newblock In \emph{CVPR}, 2014.

\bibitem[Conti et~al.(2023)Conti, Fini, Mancini, Rota, Wang, and Ricci]{conti2023vocabulary}
Alessandro Conti, Enrico Fini, Massimiliano Mancini, Paolo Rota, Yiming Wang, and Elisa Ricci.
\newblock Vocabulary-free image classification.
\newblock \emph{NeurIPS}, 2023.

\bibitem[Csurka et~al.(2024)Csurka, Hayes, Larlus, and Volpi]{csurka2024could}
Gabriela Csurka, Tyler~L Hayes, Diane Larlus, and Riccardo Volpi.
\newblock What could go wrong? discovering and describing failure modes in computer vision.
\newblock \emph{arXiv preprint arXiv:2408.04471}, 2024.

\bibitem[Dai et~al.(2023)Dai, Li, Li, Tiong, and Zhao]{dai2023instructblip}
Wenliang Dai, Junnan Li, Dongxu Li, Anthony Meng~Huat Tiong, and Junqi Zhao.
\newblock Instructblip: Towards general-purpose vision-language models with instruction tuning.
\newblock \emph{NeurIPS}, 2023.

\bibitem[Dunlap et~al.(2024)Dunlap, Zhang, Wang, Zhong, Darrell, Steinhardt, Gonzalez, and Yeung-Levy]{dunlap2024describing}
Lisa Dunlap, Yuhui Zhang, Xiaohan Wang, Ruiqi Zhong, Trevor Darrell, Jacob Steinhardt, Joseph~E Gonzalez, and Serena Yeung-Levy.
\newblock Describing differences in image sets with natural language.
\newblock In \emph{CVPR}, 2024.

\bibitem[Elo(1967)]{elo1967proposed}
Arpad~E Elo.
\newblock The proposed uscf rating system, its development, theory, and applications.
\newblock \emph{Chess life}, 22\penalty0 (8):\penalty0 242--247, 1967.

\bibitem[Eyuboglu et~al.(2022)Eyuboglu, Varma, Saab, Delbrouck, Lee-Messer, Dunnmon, Zou, and Re]{eyubogludomino}
Sabri Eyuboglu, Maya Varma, Khaled~Kamal Saab, Jean-Benoit Delbrouck, Christopher Lee-Messer, Jared Dunnmon, James Zou, and Christopher Re.
\newblock Domino: Discovering systematic errors with cross-modal embeddings.
\newblock In \emph{ICLR}, 2022.

\bibitem[Fei-Fei et~al.(2004)Fei-Fei, Fergus, and Perona]{fei2004caltech101}
Li Fei-Fei, Rob Fergus, and Pietro Perona.
\newblock Learning generative visual models from few training examples: An incremental bayesian approach tested on 101 object categories.
\newblock In \emph{CVPRW}. IEEE, 2004.

\bibitem[Geng et~al.(2020)Geng, Huang, and Chen]{geng2020open-set-survey}
Chuanxing Geng, Sheng-jun Huang, and Songcan Chen.
\newblock Recent advances in open set recognition: A survey.
\newblock \emph{IEEE TPAMI}, 43\penalty0 (10):\penalty0 3614--3631, 2020.

\bibitem[Grattafiori et~al.(2024)Grattafiori, Dubey, Jauhri, Pandey, Kadian, Al-Dahle, Letman, Mathur, Schelten, Vaughan, et~al.]{grattafiori2024llama}
Aaron Grattafiori, Abhimanyu Dubey, Abhinav Jauhri, Abhinav Pandey, Abhishek Kadian, Ahmad Al-Dahle, Aiesha Letman, Akhil Mathur, Alan Schelten, Alex Vaughan, et~al.
\newblock The llama 3 herd of models.
\newblock \emph{arXiv preprint arXiv:2407.21783}, 2024.

\bibitem[Helber et~al.(2019)Helber, Bischke, Dengel, and Borth]{helber2019eurosat}
Patrick Helber, Benjamin Bischke, Andreas Dengel, and Damian Borth.
\newblock Eurosat: A novel dataset and deep learning benchmark for land use and land cover classification.
\newblock \emph{IEEE Journal of Selected Topics in Applied Earth Observations and Remote Sensing}, 12\penalty0 (7), 2019.

\bibitem[Hsieh et~al.(2023)Hsieh, Zhang, Ma, Kembhavi, and Krishna]{hsieh2023sugarcrepe}
Cheng-Yu Hsieh, Jieyu Zhang, Zixian Ma, Aniruddha Kembhavi, and Ranjay Krishna.
\newblock Sugarcrepe: Fixing hackable benchmarks for vision-language compositionality.
\newblock \emph{NeurIPS}, 2023.

\bibitem[Huang et~al.(2023)Huang, Huang, Zhang, Tian, Feng, Zhang, Xie, Li, and Zhang]{huang2023ramplus}
Xinyu Huang, Yi-Jie Huang, Youcai Zhang, Weiwei Tian, Rui Feng, Yuejie Zhang, Yanchun Xie, Yaqian Li, and Lei Zhang.
\newblock Open-set image tagging with multi-grained text supervision.
\newblock \emph{arXiv preprint arXiv:2310.15200}, 2023.

\bibitem[Jiang et~al.(2023)Jiang, Sablayrolles, Mensch, Bamford, Chaplot, de~las Casas, Bressand, Lengyel, Lample, Saulnier, Lavaud, Lachaux, Stock, Scao, Lavril, Wang, Lacroix, and Sayed]{jiang2023mistral7b}
Albert~Q. Jiang, Alexandre Sablayrolles, Arthur Mensch, Chris Bamford, Devendra~Singh Chaplot, Diego de~las Casas, Florian Bressand, Gianna Lengyel, Guillaume Lample, Lucile Saulnier, Lélio~Renard Lavaud, Marie-Anne Lachaux, Pierre Stock, Teven~Le Scao, Thibaut Lavril, Thomas Wang, Timothée Lacroix, and William~El Sayed.
\newblock Mistral 7b, 2023.

\bibitem[Kim et~al.(2024)Kim, Mo, Kim, Lee, Lee, and Shin]{kim2024discovering}
Younghyun Kim, Sangwoo Mo, Minkyu Kim, Kyungmin Lee, Jaeho Lee, and Jinwoo Shin.
\newblock Discovering and mitigating visual biases through keyword explanation.
\newblock In \emph{CVPR}, 2024.

\bibitem[Koh et~al.(2023)Koh, Salakhutdinov, and Fried]{koh2023fromage}
Jing~Yu Koh, Ruslan Salakhutdinov, and Daniel Fried.
\newblock Grounding language models to images for multimodal inputs and outputs.
\newblock In \emph{International Conference on Machine Learning}, pages 17283--17300. PMLR, 2023.

\bibitem[Kojima et~al.(2022)Kojima, Gu, Reid, Matsuo, and Iwasawa]{kojima2022zero-shot-cot}
Takeshi Kojima, Shixiang~Shane Gu, Machel Reid, Yutaka Matsuo, and Yusuke Iwasawa.
\newblock Large language models are zero-shot reasoners.
\newblock \emph{NeurIPS}, 2022.

\bibitem[Krause et~al.(2013)Krause, Stark, Deng, and Fei-Fei]{krause20133cars}
Jonathan Krause, Michael Stark, Jia Deng, and Li Fei-Fei.
\newblock 3d object representations for fine-grained categorization.
\newblock In \emph{ICCV-WS}, 2013.

\bibitem[Lauren{\c{c}}on et~al.(2025)Lauren{\c{c}}on, Tronchon, Cord, and Sanh]{laurenccon2025idefics}
Hugo Lauren{\c{c}}on, L{\'e}o Tronchon, Matthieu Cord, and Victor Sanh.
\newblock What matters when building vision-language models?
\newblock \emph{NeurIPS}, 2025.

\bibitem[Li et~al.(2024{\natexlab{a}})Li, Ge, Ge, Wang, Wang, Zhang, and Shan]{li2024seed}
Bohao Li, Yuying Ge, Yixiao Ge, Guangzhi Wang, Rui Wang, Ruimao Zhang, and Ying Shan.
\newblock Seed-bench: Benchmarking multimodal large language models.
\newblock In \emph{CVPR}, 2024{\natexlab{a}}.

\bibitem[Li et~al.(2024{\natexlab{b}})Li, Zhang, Zhang, Guo, Zhang, Li, Zhang, Liu, and Li]{li2024llavaNext}
Bo Li, Kaichen Zhang, Hao Zhang, Dong Guo, Renrui Zhang, Feng Li, Yuanhan Zhang, Ziwei Liu, and Chunyuan Li.
\newblock Llava-next: Stronger llms supercharge multimodal capabilities in the wild.
\newblock \emph{https://llava-vl.github.io/blog/2024-05-10-llava-next-stronger-llms}, 2024{\natexlab{b}}.

\bibitem[Li et~al.(2024{\natexlab{c}})Li, Zhang, Guo, Zhang, Li, Zhang, Zhang, Zhang, Li, Liu, et~al.]{li2024llava-ov}
Bo Li, Yuanhan Zhang, Dong Guo, Renrui Zhang, Feng Li, Hao Zhang, Kaichen Zhang, Peiyuan Zhang, Yanwei Li, Ziwei Liu, et~al.
\newblock Llava-onevision: Easy visual task transfer.
\newblock \emph{arXiv preprint arXiv:2408.03326}, 2024{\natexlab{c}}.

\bibitem[Li et~al.(2023{\natexlab{a}})Li, Li, Savarese, and Hoi]{li2023blip2}
Junnan Li, Dongxu Li, Silvio Savarese, and Steven Hoi.
\newblock Blip-2: Bootstrapping language-image pre-training with frozen image encoders and large language models.
\newblock In \emph{ICML}, 2023{\natexlab{a}}.

\bibitem[Li et~al.(2024{\natexlab{d}})Li, Wang, He, Li, Wang, Liu, Wang, Xu, Chen, Luo, et~al.]{li2024mvbench}
Kunchang Li, Yali Wang, Yinan He, Yizhuo Li, Yi Wang, Yi Liu, Zun Wang, Jilan Xu, Guo Chen, Ping Luo, et~al.
\newblock Mvbench: A comprehensive multi-modal video understanding benchmark.
\newblock In \emph{CVPR}, 2024{\natexlab{d}}.

\bibitem[Li et~al.(2023{\natexlab{b}})Li, Du, Zhou, Wang, Zhao, and Wen]{li2023evaluating}
Yifan Li, Yifan Du, Kun Zhou, Jinpeng Wang, Wayne~Xin Zhao, and Ji-Rong Wen.
\newblock Evaluating object hallucination in large vision-language models.
\newblock In \emph{EMNLP}, 2023{\natexlab{b}}.

\bibitem[Liu et~al.(2023)Liu, Li, Wu, and Lee]{liu2023llava}
Haotian Liu, Chunyuan Li, Qingyang Wu, and Yong~Jae Lee.
\newblock Visual instruction tuning.
\newblock \emph{NeurIPS}, 2023.

\bibitem[Liu et~al.(2024{\natexlab{a}})Liu, Xiao, Liu, Li, Feng, Yang, and Wang]{liu2024revisiting}
Huan Liu, Lingyu Xiao, Jiangjiang Liu, Xiaofan Li, Ze Feng, Sen Yang, and Jingdong Wang.
\newblock Revisiting mllms: An in-depth analysis of image classification abilities.
\newblock \emph{arXiv preprint arXiv:2412.16418}, 2024{\natexlab{a}}.

\bibitem[Liu et~al.(2024{\natexlab{b}})Liu, Duan, Zhang, Li, Zhang, Zhao, Yuan, Wang, He, Liu, et~al.]{liu2024mmbench}
Yuan Liu, Haodong Duan, Yuanhan Zhang, Bo Li, Songyang Zhang, Wangbo Zhao, Yike Yuan, Jiaqi Wang, Conghui He, Ziwei Liu, et~al.
\newblock Mmbench: Is your multi-modal model an all-around player?
\newblock In \emph{ECCV}, 2024{\natexlab{b}}.

\bibitem[Maji et~al.(2013)Maji, Rahtu, Kannala, Blaschko, and Vedaldi]{maji2013aircraft}
Subhransu Maji, Esa Rahtu, Juho Kannala, Matthew Blaschko, and Andrea Vedaldi.
\newblock Fine-grained visual classification of aircraft.
\newblock \emph{arXiv preprint arXiv:1306.5151}, 2013.

\bibitem[Moayeri et~al.(2022)Moayeri, Singla, and Feizi]{moayeri2022hard}
Mazda Moayeri, Sahil Singla, and Soheil Feizi.
\newblock Hard imagenet: Segmentations for objects with strong spurious cues.
\newblock \emph{NeurIPS}, 2022.

\bibitem[Nilsback and Zisserman(2008)]{nilsback2008flowers}
Maria-Elena Nilsback and Andrew Zisserman.
\newblock Automated flower classification over a large number of classes.
\newblock In \emph{Indian conference on computer vision, graphics \& image processing}. IEEE, 2008.

\bibitem[Parkhi et~al.(2012)Parkhi, Vedaldi, Zisserman, and Jawahar]{parkhi2012pets}
Omkar~M Parkhi, Andrea Vedaldi, Andrew Zisserman, and CV Jawahar.
\newblock Cats and dogs.
\newblock In \emph{CVPR}, 2012.

\bibitem[Peychev et~al.(2023)Peychev, M{\"u}ller, Fischer, and Vechev]{peychev2023automated}
Momchil Peychev, Mark M{\"u}ller, Marc Fischer, and Martin Vechev.
\newblock Automated classification of model errors on imagenet.
\newblock \emph{NeurIPS}, 2023.

\bibitem[Radford et~al.(2021)Radford, Kim, Hallacy, Ramesh, Goh, Agarwal, Sastry, Askell, Mishkin, Clark, et~al.]{radford2021clip}
Alec Radford, Jong~Wook Kim, Chris Hallacy, Aditya Ramesh, Gabriel Goh, Sandhini Agarwal, Girish Sastry, Amanda Askell, Pamela Mishkin, Jack Clark, et~al.
\newblock Learning transferable visual models from natural language supervision.
\newblock In \emph{ICML}, 2021.

\bibitem[Reimers and Gurevych(2019)]{reimers2019sentence}
Nils Reimers and Iryna Gurevych.
\newblock Sentence-bert: Sentence embeddings using siamese bert-networks.
\newblock In \emph{EMNLP-IJCNLP}, 2019.

\bibitem[Russakovsky et~al.(2015)Russakovsky, Deng, Su, Krause, Satheesh, Ma, Huang, Karpathy, Khosla, Bernstein, et~al.]{russakovsky2015imagenet}
Olga Russakovsky, Jia Deng, Hao Su, Jonathan Krause, Sanjeev Satheesh, Sean Ma, Zhiheng Huang, Andrej Karpathy, Aditya Khosla, Michael Bernstein, et~al.
\newblock Imagenet large scale visual recognition challenge.
\newblock \emph{IJCV}, 115:\penalty0 211--252, 2015.

\bibitem[Shu et~al.(2022)Shu, Nie, Huang, Yu, Goldstein, Anandkumar, and Xiao]{shu2022tpt}
Manli Shu, Weili Nie, De-An Huang, Zhiding Yu, Tom Goldstein, Anima Anandkumar, and Chaowei Xiao.
\newblock Test-time prompt tuning for zero-shot generalization in vision-language models.
\newblock \emph{NeurIPS}, 2022.

\bibitem[Singh et~al.(2022)Singh, Hu, Goswami, Couairon, Galuba, Rohrbach, and Kiela]{singh2022flava}
Amanpreet Singh, Ronghang Hu, Vedanuj Goswami, Guillaume Couairon, Wojciech Galuba, Marcus Rohrbach, and Douwe Kiela.
\newblock Flava: A foundational language and vision alignment model.
\newblock In \emph{CVPR}, 2022.

\bibitem[Singla and Feizi(2022)]{singlasalient}
Sahil Singla and Soheil Feizi.
\newblock Salient imagenet: How to discover spurious features in deep learning?
\newblock In \emph{ICLR}, 2022.

\bibitem[Singla et~al.(2021)Singla, Nushi, Shah, Kamar, and Horvitz]{singla2021understanding}
Sahil Singla, Besmira Nushi, Shital Shah, Ece Kamar, and Eric Horvitz.
\newblock Understanding failures of deep networks via robust feature extraction.
\newblock In \emph{CVPR}, 2021.

\bibitem[Soomro et~al.(2012)Soomro, Zamir, and Shah]{soomro2012ucf101}
Khurram Soomro, Amir~Roshan Zamir, and Mubarak Shah.
\newblock Ucf101: A dataset of 101 human actions classes from videos in the wild.
\newblock \emph{arXiv preprint arXiv:1212.0402}, 2012.

\bibitem[Thawakar et~al.(2025)Thawakar, Dissanayake, More, Thawkar, Heakl, Ahsan, Li, Zumri, Lahoud, Anwer, et~al.]{thawakar2025llamav}
Omkar Thawakar, Dinura Dissanayake, Ketan More, Ritesh Thawkar, Ahmed Heakl, Noor Ahsan, Yuhao Li, Mohammed Zumri, Jean Lahoud, Rao~Muhammad Anwer, et~al.
\newblock Llamav-o1: Rethinking step-by-step visual reasoning in llms.
\newblock \emph{arXiv preprint arXiv:2501.06186}, 2025.

\bibitem[Touvron et~al.(2023)Touvron, Lavril, Izacard, Martinet, Lachaux, Lacroix, Rozi{\`e}re, Goyal, Hambro, Azhar, et~al.]{touvron2023llama}
Hugo Touvron, Thibaut Lavril, Gautier Izacard, Xavier Martinet, Marie-Anne Lachaux, Timoth{\'e}e Lacroix, Baptiste Rozi{\`e}re, Naman Goyal, Eric Hambro, Faisal Azhar, et~al.
\newblock Llama: Open and efficient foundation language models.
\newblock \emph{arXiv preprint arXiv:2302.13971}, 2023.

\bibitem[Vasudevan et~al.(2022)Vasudevan, Caine, Gontijo~Lopes, Fridovich-Keil, and Roelofs]{vasudevan2022does}
Vijay Vasudevan, Benjamin Caine, Raphael Gontijo~Lopes, Sara Fridovich-Keil, and Rebecca Roelofs.
\newblock When does dough become a bagel? analyzing the remaining mistakes on imagenet.
\newblock \emph{NeurIPS}, 2022.

\bibitem[Wang et~al.(2024)Wang, Bai, Tan, Wang, Fan, Bai, Chen, Liu, Wang, Ge, et~al.]{wang2024qwen2}
Peng Wang, Shuai Bai, Sinan Tan, Shijie Wang, Zhihao Fan, Jinze Bai, Keqin Chen, Xuejing Liu, Jialin Wang, Wenbin Ge, et~al.
\newblock Qwen2-vl: Enhancing vision-language model's perception of the world at any resolution.
\newblock \emph{arXiv preprint arXiv:2409.12191}, 2024.

\bibitem[Wei et~al.(2022)Wei, Wang, Schuurmans, Bosma, Xia, Chi, Le, Zhou, et~al.]{wei2022cot}
Jason Wei, Xuezhi Wang, Dale Schuurmans, Maarten Bosma, Fei Xia, Ed Chi, Quoc~V Le, Denny Zhou, et~al.
\newblock Chain-of-thought prompting elicits reasoning in large language models.
\newblock \emph{NeurIPS}, 2022.

\bibitem[Xiao et~al.(2010)Xiao, Hays, Ehinger, Oliva, and Torralba]{xiao2010sun}
Jianxiong Xiao, James Hays, Krista~A Ehinger, Aude Oliva, and Antonio Torralba.
\newblock Sun database: Large-scale scene recognition from abbey to zoo.
\newblock In \emph{CVPR}, 2010.

\bibitem[Xu et~al.(2024)Xu, Jin, Hao, Song, Sun, and Yuan]{xu2024llavacot}
Guowei Xu, Peng Jin, Li Hao, Yibing Song, Lichao Sun, and Li Yuan.
\newblock Llava-o1: Let vision language models reason step-by-step.
\newblock \emph{arXiv preprint arXiv:2411.10440}, 2024.

\bibitem[Yenamandra et~al.(2023)Yenamandra, Ramesh, Prabhu, and Hoffman]{yenamandra2023facts}
Sriram Yenamandra, Pratik Ramesh, Viraj Prabhu, and Judy Hoffman.
\newblock Facts: First amplify correlations and then slice to discover bias.
\newblock In \emph{ICCV}, 2023.

\bibitem[Yu et~al.(2024)Yu, Shen, and Chen]{yu2024towards}
Qihang Yu, Xiaohui Shen, and Liang-Chieh Chen.
\newblock Towards open-ended visual recognition with large language models.
\newblock In \emph{ECCV}, 2024.

\bibitem[Yue et~al.(2024)Yue, Chen, Geiping, Li, Goldstein, and Lim]{yue2024object}
Kaiyu Yue, Bor-Chun Chen, Jonas Geiping, Hengduo Li, Tom Goldstein, and Ser-Nam Lim.
\newblock Object recognition as next token prediction.
\newblock In \emph{CVPR}, 2024.

\bibitem[Yuksekgonul et~al.(2023)Yuksekgonul, Bianchi, Kalluri, Jurafsky, and Zou]{yuksekgonuland}
Mert Yuksekgonul, Federico Bianchi, Pratyusha Kalluri, Dan Jurafsky, and James Zou.
\newblock When and why vision-language models behave like bags-of-words, and what to do about it?
\newblock In \emph{ICLR}, 2023.

\bibitem[Zhai et~al.(2023)Zhai, Mustafa, Kolesnikov, and Beyer]{zhai2023siglip}
Xiaohua Zhai, Basil Mustafa, Alexander Kolesnikov, and Lucas Beyer.
\newblock Sigmoid loss for language image pre-training.
\newblock In \emph{ICCV}, 2023.

\bibitem[Zhang et~al.(2024)Zhang, Unell, Wang, Ghosh, Su, Schmidt, and Yeung-Levy]{zhang2024visually}
Yuhui Zhang, Alyssa Unell, Xiaohan Wang, Dhruba Ghosh, Yuchang Su, Ludwig Schmidt, and Serena Yeung-Levy.
\newblock Why are visually-grounded language models bad at image classification?
\newblock \emph{NeurIPS}, 2024.

\bibitem[Zheng et~al.(2023{\natexlab{a}})Zheng, He, and Wang]{zheng2023minigpt5}
Kaizhi Zheng, Xuehai He, and Xin~Eric Wang.
\newblock Minigpt-5: Interleaved vision-and-language generation via generative vokens.
\newblock \emph{arXiv preprint arXiv:2310.02239}, 2023{\natexlab{a}}.

\bibitem[Zheng et~al.(2023{\natexlab{b}})Zheng, Chiang, Sheng, Zhuang, Wu, Zhuang, Lin, Li, Li, Xing, et~al.]{zheng2023judging}
Lianmin Zheng, Wei-Lin Chiang, Ying Sheng, Siyuan Zhuang, Zhanghao Wu, Yonghao Zhuang, Zi Lin, Zhuohan Li, Dacheng Li, Eric Xing, et~al.
\newblock Judging llm-as-a-judge with mt-bench and chatbot arena.
\newblock \emph{NeurIPS}, 2023{\natexlab{b}}.

\bibitem[Zhou et~al.(2022)Zhou, Yang, Loy, and Liu]{zhou2022learning}
Kaiyang Zhou, Jingkang Yang, Chen~Change Loy, and Ziwei Liu.
\newblock Learning to prompt for vision-language models.
\newblock \emph{IJCV}, 130\penalty0 (9):\penalty0 2337--2348, 2022.

\bibitem[Zhou et~al.(2024)Zhou, Cui, Yoon, Zhang, Deng, Finn, Bansal, and Yao]{zhou2023analyzing}
Yiyang Zhou, Chenhang Cui, Jaehong Yoon, Linjun Zhang, Zhun Deng, Chelsea Finn, Mohit Bansal, and Huaxiu Yao.
\newblock Analyzing and mitigating object hallucination in large vision-language models.
\newblock \emph{ICLR}, 2024.

\bibitem[Zhu et~al.(2024)Zhu, Chen, Shen, Li, and Elhoseiny]{zhu2024minigpt4}
Deyao Zhu, Jun Chen, Xiaoqian Shen, Xiang Li, and Mohamed Elhoseiny.
\newblock Minigpt-4: Enhancing vision-language understanding with advanced large language models.
\newblock In \emph{ICLR}, 2024.

\end{thebibliography}
